\RequirePackage{rotating}

\documentclass[reqno]{hdsr}

\graphicspath{{figs/}}

\usepackage[normalem]{ulem} 

\usepackage{scrextend}
\usepackage{graphicx}
\usepackage{rotating}
\usepackage{array}
\usepackage{algorithm}
\usepackage{algorithmic}
\usepackage[labelfont=bf]{caption}
\usepackage{xcolor,colortbl}
\usepackage{setspace}
\usepackage{ulem}
\usepackage{cancel}

\usepackage{bm}
\usepackage{amsmath}
\usepackage{amssymb}
\usepackage{amsfonts}
\usepackage{mathtools}

\newcommand{\cbr}[1]{\left\{#1\right\}}

\renewcommand{\algorithmicrequire}{\textbf{Input:}}
\renewcommand{\algorithmicensure}{\textbf{Output:}}

\renewcommand{\E}{\mathbb{E}}
\newcommand{\KLD}{\text{KL}}
\newcommand{\ENT}{\text{H}}
\newcommand{\BENT}{\mathbb{H}}
\newcommand{\BDist}{\mathbb{D}}

\newcommand{\x}{\bm{x}}
\newcommand{\y}{\bm{y}}

\newcommand{\bt}{\bm{t}}
\newcommand{\xspace}{\mathcal{X}}
\newcommand{\yspace}{\mathcal{Y}}

\newcommand{\tspace}{\mathcal{T}}

\newcommand{\btheta}{\bm{\theta}}
\newcommand{\bphi}{\bm{\phi}}
\newcommand{\bxi}{\bm{\xi}}

\newcommand{\dataset}{\mathcal{D}}
\newcommand{\loss}{\mathcal{L}}
\newcommand{\indicator}{\mathbb{I}}
\newcommand{\pdata}{p_{d}}

\newcolumntype{L}[1]{>{\raggedright\arraybackslash}p{#1}}
\newcolumntype{R}[1]{>{\raggedleft\arraybackslash}p{#1}}
\newcolumntype{C}[1]{>{\centering\let\newline\\\arraybackslash\hspace{0pt}}m{#1}}
\newcolumntype{?}{!{\vrule width 1pt}}
\makeatletter
\newcommand{\thickhline}{%
    \noalign {\ifnum 0=`}\fi \hrule height 1pt
    \futurelet \reserved@a \@xhline
}
\newcolumntype{"}{@{\hskip\tabcolsep\vrule width 1pt\hskip\tabcolsep}}
\makeatother

\definecolor{Gray}{gray}{0.94}
\newcolumntype{g}{>{\columncolor{Gray}}l}

\DeclareMathOperator*{\argmax}{\mathrm{argmax}}
\DeclareMathOperator*{\argmin}{\mathrm{argmin}}

\newcommand{\intset}[1]{\cbr{1..n}}

\usepackage{lscape} 

\definecolor{dark-red}{rgb}{0.4,0.15,0.15}
\definecolor{dark-blue}{rgb}{0.15,0.15,0.4}
\definecolor{medium-blue}{rgb}{0,0,0.5}
\hypersetup{
   colorlinks, linkcolor={dark-blue},
   citecolor={dark-blue}, urlcolor={medium-blue}
}

\usepackage{booktabs}
\usepackage{multirow}
\usepackage{comment}
\usepackage{parskip}
\usepackage{gensymb}

\numberwithin{equation}{section}

\newcommand{\hztlast}[1]{\textcolor{red}{}}

\begin{document}

\newgeometry{bottom=1.5in}

\volumeheader{4}{4}{10.1162}

\begin{center}

  \title{Toward a `Standard Model' of Machine Learning}
  \maketitle
 
  \thispagestyle{empty}
  
  \vspace*{.2in}

  \begin{tabular}{cc}
    Zhiting Hu\upstairs{\affilone,*},~~ Eric P. Xing\upstairs{\affiltwo,\affilthree,\affilfour,**}
   \\[0.25ex]
   {\small \upstairs{\affilone} Halıcıoğlu Data Science Institute, University of California San Diego, San Diego, USA} \\
   {\small \upstairs{\affiltwo} Machine Learning Department, Carnegie Mellon University, Pittsburgh, USA} \\
   {\small \upstairs{\affilfour} Mohamed bin Zayed University of Artificial Intelligence, Abu Dhabi, UAE}\\
  {\small \upstairs{\affilthree} Petuum Inc., Pittsburgh, USA}
  \end{tabular}
  
  \emails{
    \upstairs{*}zhh019@ucsd.edu, \upstairs{**}epxing@cs.cmu.edu 
    }
  \vspace*{0.4in}
 
\begin{abstract}
Machine learning (ML) is about computational methods that enable machines to learn concepts from experience. In handling a wide variety of experience ranging from data instances, knowledge, constraints, to rewards, adversaries, and lifelong interaction in an ever-growing spectrum of tasks, contemporary ML/AI (artificial intelligence) research has resulted in a multitude of learning paradigms and methodologies. Despite the continual progresses on all different fronts, the disparate narrowly focused methods also make standardized, composable, and reusable development of ML approaches difficult, and preclude the opportunity to build AI agents that panoramically learn from all types of experience. This article presents a standardized ML formalism, in particular a `standard equation' of the learning objective, that offers a unifying understanding of many important ML algorithms in the supervised, unsupervised, knowledge-constrained, reinforcement, adversarial, and online learning paradigms, respectively---those diverse algorithms are encompassed as special cases due to different choices of modeling components. The framework also provides guidance for mechanical design of new ML approaches and serves as a promising vehicle toward {\it panoramic} machine learning with all experience.
\end{abstract}
\end{center}

\vspace*{0.15in}
\hspace{10pt}
  
\copyrightnotice

\section*{Media Summary}
Humans learn from a range of experience, what about a computer? 
The past decades of AI and Machine Learning (ML) research has resulted in a multitude of paradigms and algorithms each specialized to train ML models with a certain type of information and experience in a certain type of problem. While pushing the field forward rapidly, the bewildering and ever-growing variety of paradigms and algorithms also makes it extremely difficult to master existing ML techniques and to develop universal, repeatable, and reusable computer programs that can simultaneously learn from diverse experience in the real world. It is a constant desire and aspiration to search for a standardized ML formalism that unifies the distinct learning principles, much like the Standard Model in physics, to gain a more holistic understanding of the diverse paradigms and algorithms, lay out a blueprint permitting fuller and more systematic exploration in the design and analysis of new algorithms, and eventually serves as a vehicle toward panoramic machine learning capable of integrating all available information (data, knowledge, constraints, reward, adversary, etc.) in learning and thus applicable to all problems. This work presents an attempt toward this end. In particular, we establish a standard equation of the learning objective, which subsumes many of the known algorithms as special cases and offers guiding principles of designing new more powerful learning algorithms in a mechanical and composable way.


\section{Introduction}
\label{sec1}

Human learning has the hallmark of learning concepts from diverse sources of information. Take the example of learning a language. Humans can benefit from various experience---by observing examples through reading and hearing, studying abstract definitions and grammar, making mistakes and getting correction from teachers, interacting with others and observing implicit feedback, and so on. Knowledge of a prior language can also  accelerate the acquisition of a new one. How can we build artificial intelligence (AI) agents that are similarly capable of learning from all types of experience? We refer to the capability of flexibly integrating all available experience in learning as \emph{panoramic learning}.

In handling different experience ranging from data instances, knowledge, constraints, to rewards, adversaries, and lifelong interplay in an ever-growing spectrum of tasks, contemporary ML and AI research has resulted in a large multitude of learning paradigms (e.g., supervised, unsupervised, active, reinforcement, adversarial learning), models, optimization techniques, not mentioning countless approximation heuristics and tuning tricks, plus combinations of all the above. While pushing the field forward rapidly, these results also make mastering existing ML techniques very difficult, and fall short of reusable, repeatable, and composable development of ML approaches to diverse problems with distinct available experience. 


Those fundamental challenges call for a standardized ML formalism that offers a principled framework for understanding, unifying, and generalizing current major paradigms of learning algorithms, and for mechanical design of new approaches for integrating any useful experience in learning. 
The power of standardized theory is perhaps best demonstrated in physics, which has a long history of pursuing symmetry and simplicity of its principles: exemplified by the famed Maxwell's equations in the 1800s that reduced various principles of electricity and magnetism into a single electromagnetic theory, followed by General Relativity in the 1910s and the Standard Model in the 1970s, physicists describe the world best by unifying and reducing different theories to a standardized one. 
Likewise, it is a constant quest in the field of ML to establish a `Standard Model' \citep{langley1989toward,domingos2015master}, that gives a holistic view of the broad learning principles, lays out a blueprint permitting fuller and more systematic exploration in the design and analysis of new algorithms, and eventually serves as a vehicle toward panoramic learning that integrates all available sources of experience. 




This paper presents an attempt toward this end. In particular, our principal focus is on the \emph{learning objective} that drives the model training given the experience and thus often lies at the core for 
designing new algorithms, understanding learning properties, and validating outcomes. 
We investigate the underlying connections between a range of seemingly distinct ML paradigms. Each of these paradigms has made particular assumptions on the form of experience available.
For example, the present most popular {\it supervised learning} relies on collections of data instances, often applying a maximum likelihood objective solved with simple gradient descent. Maximum likelihood based {\it unsupervised learning} instead can invoke different solvers, such as expectation-maximization (EM), variational inference, and wake-sleep in training for varied degree of approximation to the problem. 
{\it Active learning} \citep{settles2012active} manages data instances which, instead of being given all at once, are adaptively selected. {\it Reinforcement learning} \citep{sutton2017reinforcement} makes use of feedback obtained via interaction with the environment. {\it Knowledge-constrained learning} like posterior regularization \citep{ganchev2010posterior,zhu2014bayesian} incorporates structures, knowledge, and rules expressed as constraints.  {\it Generative adversarial learning} \citep{goodfellow2014generative} leverages a companion model called discriminator to guide training of the model of interest. 

In light of these results, we present a standard equation (SE) of the objective function. The SE formulates a rather broad design space of learning algorithms. We show that many of the well-known algorithms of the above paradigms are all instantiations of the general formulation. More concretely, the SE, based on the maximum entropy and variational principles, consists of three principled terms, including the {\it experience} term that offers a unified language to express arbitrary relevant information to supervise the learning, the {\it divergence} term that measures the fitness of the target model to be learned, and the {\it uncertainty} term that regularizes the complexity of the system. The single succinct formula re-derives the objective functions of a large diversity of learning algorithms, reducing them to different choices of the components. The formulation thus shed new light on the fundamental relationships between the diverse algorithms that were each originally designed to deal with a specific type of experience. 

The modularity and generality of the framework is particularly appealing not only from the theoretical point of view, but also because it offers guiding principles for designing algorithmic approaches to new problems in a mechanical way. Specifically, the SE by its nature allows combining together all different experience to learn a model of interest. Designing a problem solution boils down to choosing {\it what} experience to use depending on the problem structure and available resources, without worrying too much about {\it how} to use the experience in the training. 
Besides, the standardized ML perspective also highlights that many learning problems in different research areas are essentially the same and just correspond to different specifications of the SE components. This enables us to systematically repurpose successful techniques in one area to solve problems in another.

The remainder of the article is organized as follows. Section~\ref{sec:background} gives an overview of relevant learning and inference techniques as a prelude of the standardized framework. Section~\ref{sec:se} presents the standard equation as a general formulation of the objective function in learning algorithms. The subsequent two sections discuss different choices of two of the key components in the standard equation, respectively, illustrating that many existing methods are special cases of the formulation: Section~\ref{sec:experience} is devoted to discussion of the experience function and Section~\ref{sec:divergence} focuses on the divergence function. Section~\ref{sec:advanced-experiences} discusses an extended view of the standard equation in dynamic environments.
Section~\ref{sec:opt} focuses on the optimization algorithms for solving the standard equation objective. Section~\ref{sec:target-model} discusses the diverse types of target models. Section~\ref{sec:app} discusses the utility of the standardized formalism for mechanical design of panoramic learning approaches. Section~\ref{sec:related} reviews related work. Section~\ref{sec:future} concludes the article with discussion of future directions---in particular, 
we discuss the broader aspects of ML not covered in the present work (e.g., more advanced learning such as continual learning in complex evolving environments, 
theoretical analysis of learnability, generalization and complexity, and automated algorithm composition) and how their unified characterization based on or inspired by the current framework could potentially lead toward a full `Standard Model' of ML and a turnkey approach to panoramic learning with all types of experience.

\section{Preliminaries: The Maximum Entropy View of Learning and Inference}\label{sec:background}

Depending on the nature of the task (e.g., classification or regression), data (e.g., labeled or unlabeled), information scope (e.g., with or without latent variables), and form of domain knowledge (e.g., prior distributions or parameter constraints), and so on, different learning paradigms with often complementary (but not necessarily easy to combine) advantages have been developed for different needs. For example, the paradigms built on the maximum likelihood principles, Bayesian theories, variational calculus, and Monte Carlo simulation have led to much of the foundation underlying a wide spectrum of probabilistic graphical models, exact/approximate inference algorithms, and even probabilistic logic programs suitable for probabilistic inference and parameter estimations in multivariate, structured, and fully or partially observed domains, while the paradigms built on convex optimization, duality theory, regularization, and risk minimization 
have led to much of the foundation underlying algorithms such as support vector machine (SVM), boosting, sparse learning, structure learning, and so on. Historically, there have been numerous efforts in establishing a unified machine learning framework that can bridge these complementary paradigms so that advantages in model design, solver efficiency, side-information incorporation, and theoretical guarantees can be translated across paradigms. 
As a prelude of our presentation of the `standard equation' framework toward this goal, here we begin with a recapitulation of the maximum entropy view of statistical learning. By naturally marrying the probabilistic frameworks with the optimization-theoretic frameworks, the maximum entropy viewpoint had played an important historical role in offering the same lens to understanding several popular methodologies such as maximum likelihood learning, Bayesian inference, and large margin learning.

\subsection{Maximum Likelihood Estimation (MLE)}
We start with the maximum entropy perspective of the maximum likelihood learning.

\subsubsection{\textbf{Supervised MLE}}\label{sec:background:sup-mle}
We consider an arbitrary probabilistic model (e.g., a neural network or probabilistic graphical model for, say, language generation) with parameters $\btheta\in\bm{\Theta}$ to be learned. Let $p_\theta(\x)\in\mathcal{P}(\xspace)$ denote the distribution defined by the model, where $\xspace$ is the data space (e.g., all language text) and $\mathcal{P}(\xspace)$ denotes the set of all probability distributions on $\xspace$. Given a set of independent and identically distributed (i.i.d.) data examples $\mathcal{D}=\{ \x^* \in \xspace \}$, 
the most common method for estimating the parameters $\btheta$ is perhaps maximum likelihood estimation (MLE). 
MLE learns the model by minimizing the negative log-likelihood:
\begin{equation}
\begin{split}
    \min_{\btheta} - \E_{\x^* \sim \mathcal{D}}\left[ \log p_\theta(\x^*) \right].
\end{split}
\label{eq:mle-ori}
\end{equation}
MLE is known to be intimately related to the \emph{maximum entropy} principle~\citep{jaynes1957information}.
In particular, when the model $p_\theta(\x)$ is in the \emph{exponential family} of the form:
\begin{equation}
\begin{split}
    p_\theta(\x) = \exp\left\{ \btheta \cdot T(\x) \right\} / Z(\btheta),
\end{split}
\label{eq:exp-family}
\end{equation}
where $T(\x)$ is the sufficient statistics of data $\x$ and
$Z(\btheta)=\sum_{\x\in\xspace} \exp\{\btheta\cdot T(\x)\}$ is the normalization factor, it is shown that MLE is the convex dual of maximum entropy estimation. 

In a maximum entropy formulation, rather than assuming a specific parametric from of the target model distribution, denoted as $p(\x)$, we instead impose constraints on the model distribution. Specifically, in the supervised setting, the constraints require the expectation of the features $T(\x)$ to be equal to the empirical expectation:
\begin{equation}
\begin{split}
    \E_{p}\left[ T(\x) \right] = \E_{\x^*\sim \mathcal{D}}\left[ T(\x^*) \right].
\end{split}
\label{eq:mle-maxent-constraint}
\end{equation}
In general, there exist many distributions $p\in\mathcal{P}(\xspace)$ that satisfy the constraint. The principle of maximum entropy resolves the ambiguity by choosing the distribution such that its Shannon entropy, $\ENT(p):=-\E_p[\log p(\x)]$, is maximized. 
Following this principle, in the supervised setting, we thus have the specific constrained optimization problem:
\begin{equation}
\begin{split}
    \max_{p(\x)}&~~ \ENT\left( p(\x) \right) \\[3pt]
    s.t.&~~ \E_{p}\left[ T(\x) \right] = \E_{\x^*\sim \mathcal{D}}\left[ T(\x^*) \right] \\[2pt]
    &~~ p(\x) \in \mathcal{P}(\xspace).
\end{split}
\label{eq:mle-maxent}
\end{equation}

The problem can be solved with the Lagrangian method. Specifically, we write the Lagrangian:
\begin{equation}
\begin{split}
    \mathcal{L}(p, \btheta, \mu) = \ENT( p(\x) ) - \btheta\cdot\left( \E_p\left[ T(\x) \right] - \E_{\x^*\sim\mathcal{D}}\left[ T(\x^*) \right] \right) - \mu\left( \sum\nolimits_{\x} p(\x) - 1 \right),
\end{split}
\label{eq:lagrangian}
\end{equation}
where $\btheta$ and $\mu$ are Lagrangian multipliers. 
Setting the derivative w.r.t. $p$ and $\mu$ to equal zero implies that $p$ must have the same form as in Equation~\ref{eq:exp-family}:
\begin{equation}
\begin{split}
    p(\x) = \exp\left\{ \btheta \cdot T(\x) \right\} / Z(\btheta),
\end{split}
\label{eq:exp-family-2}
\end{equation}
where we see the parameters $\btheta$ in the exponential family parameterization are the Lagrangian multipliers that enforce the constraints.
Plugging the solution back into the Lagrangian, we obtain:
\begin{equation}
\begin{split}
    \mathcal{L}(\btheta) = \E_{\x^*\sim\mathcal{D}}\left[ \btheta\cdot T(\x^*) \right] - \log Z(\btheta),
\end{split}
\end{equation}
which is simply the negative of the MLE objective in Equation~\ref{eq:mle-ori}. 

Thus maximum entropy is dual to maximum likelihood. It provides an alternative view of the problem of fitting a model into data, where the data instances in the training set are treated as constraints, and the learning problem is treated as a constrained optimization problem. This optimization-theoretic view of learning will be revisited repeatedly in the sequel to allow extending machine learning under all experience of which data instances is just a special case.

\subsubsection{\textbf{Unsupervised MLE}}\label{sec:unsup-mle}

Similar to the MLE framework for supervised learning, unsupervised learning via MLE can also be reformulated as a constraint optimization problem with entropy maximization. Consider learning a multivariate model with latent variables, where each data instance is partitioned into observed variables $\x\in\xspace$ and latent variables $\y\in\yspace$. For example, in the problem of image clustering, $\x\in\mathbb{R}^{d}$ is the observed image of $d$ pixels and $\y\in\{1,\dots,K\}$ is the unobserved cluster indicator (where $K$ is the number of clusters). 
The goal is to learn a model $p_\theta(\x,\y)$ that captures the joint distribution of $\x$ and $\y$. Since $\y$ is unobserved, we minimize the negative log-likelihood with $\y$ marginalized out:
\begin{equation}
\begin{split}
    \min_{\btheta} - \E_{\x^*\sim\mathcal{D}}\left[ \log \sum_{\y\in\yspace} p_\theta(\x^*,\y) \right].
\end{split}
\label{eq:marginal-mle}
\end{equation}
Direct optimization of the marginal log-likelihood is typically intractable due to the summation over $\y$. Earlier work thus developed different solvers with varying levels of approximations.

It can be shown that the intractable negative log-likelihood above can be upper bounded by a more tractable term known as the {\it variational free energy}~\citep{neal1998view}. Let $q(\y|\x)$ represent an arbitrary auxiliary distribution acting as a surrogate of the true posterior $p(\y|\x)$, which is known as a variational distribution.
Then, for each instance $\x^*\in\dataset$, we have:
\begin{equation}
\begin{aligned}
    - \log \sum_{\y} p_{\theta}(\x^*, \y) &= - \E_{q(\y|\x^*)}\left[\log \frac{p_\theta(\x^*,\y)}{q(\y|\x^*)} \right] - \KLD\left( q(\y|\x^*) \| p_\theta(\y|\x^*) \right) \\
    &\leq - \E_{q(\y|\x^*)}\left[\log \frac{p_\theta(\x^*,\y)}{q(\y|\x^*)} \right] \\
    &= - \ENT\left(q(\y|\x^*)\right) - \E_{q(\y|\x^*)}\left[ \log p_\theta(\x^*, \y)  \right] := \loss(q,\btheta),
\end{aligned}
\label{eq:em-free-energy}
\end{equation}
where the inequality holds because KL divergence is always nonnegative. The free energy upper bound contains two terms: the first one is the entropy of the variational distribution, which captures the intrinsic randomness (i.e., amount of information carried by an auxiliary distribution); the second term, now written as $-\E_{q(\y|\x^*)\tilde{p}_d(\x^*)}\left[ \log p_\theta(\x^*, \y)  \right]$, by taking into account the empirical distribution $\tilde{p}_d$ from which the instance $\x^*$ is drawn, is the \emph{cross entropy} between the distributions $q(\y|\x^*)\tilde{p}_d(\x^*)$ and $p_\theta(\x^*, \y)$, driving the two to be close and thereby allowing $q$ to approximate $p$. 

The popular expectation maximization (EM) algorithm for unsupervised learning via MLE can be interpreted as minimizing the variational free energy~\citep{neal1998view}. In fact, as we discuss subsequently, popular heuristics such as the variational EM and the wake-sleep algorithms, are approximations to the EM algorithm by introducing approximating realizations to either the free energy objective function $\loss$ or to the solution space of the variational distribution $q$.

{\bf Expectation Maximization (EM).} 
The most common approach to learning with unlabeled data or partially observed multivariate models is perhaps the EM algorithm \citep{dempster1977maximum}. With the use of the variational free energy as a surrogate objective to the original marginal likelihood as in Equation~\ref{eq:em-free-energy},
EM can be also understood as an alternating minimization algorithm, where $\loss(q,\btheta)$ is minimized with regard to $q$ and $\btheta$ in two stages, respectively. At each iteration $n$, the expectation (E) step maximizes $\loss(q,\btheta^{(n)})$ w.r.t. $q$. From Equation~\ref{eq:em-free-energy}, this is achieved by setting $q$ to the current true posterior:
\begin{equation}
\begin{aligned}
    \text{E-step:}\quad q^{(n+1)}(\y|\x^*) = p_{\theta^{(n)}}(\y|\x^*), \qquad\ \ \ 
\end{aligned}
\label{eq:em-e}
\end{equation}
so that the KL divergence vanishes and the upper bound is tight. In the subsequent maximization (M) step, $\loss(q^{(n+1)},\btheta)$ is minimized w.r.t. $\btheta$:
\begin{equation}
\begin{aligned}
    \text{M-step:}\quad \max_{\btheta} \E_{q^{(n+1)}(\y|\x^*)}\left[ \log p_\theta(\x^*, \y)  \right],
\end{aligned}
\label{eq:em-m}
\end{equation}
which is to maximize the expected complete data log-likelihood. The EM algorithm has an appealing property that it monotonically decreases the negative marginal log-likelihood over iterations.
To see this, notice that after the E-step the upper bound $\loss(q^{(n+1)},\btheta^{(n)})$ is equal to the negative marginal log-likelihood, and the M-step further decreases the upper bound (and thus the negative marginal log-likelihood).


{\bf Variational EM.}
When the model $p_\theta(\x, \y)$ is complex (e.g., a neural network or a multilayer graphical model), directly working with the true posterior in the E-step becomes intractable. Variational EM overcomes the difficulty with approximations. It considers a restricted family $\mathcal{Q}'$ of the variational distribution $q(\y)$ such that optimization w.r.t. $q$ within the family is tractable:
\begin{equation}
\begin{split}
    \text{Variational E-step:}\quad \min_{q\in\mathcal{Q}'} \loss(q, \btheta^{(t)}).
\end{split}
\label{eq:vem-e}
\end{equation}
A common way to restrict the $q$ family is the \emph{mean-field methods}, which partition the components of $\y$ into sub-groups $\y=(\y_1, \dots, \y_M)$ and assume that $q$ factorizes w.r.t. the groups: $q(\y)=\prod_{i=1}^{M}q_i(\y_i)$. 
The variational principle summarized in~\citep{wainwright2008graphical} gives a more principled interpretation of the mean-field and other approximation methods. In particular,
in the case where $p_\theta(\x,\y)$ is an exponential family distribution with sufficient statistics $T(\x,\y)$, the exact E-step (Equation~\ref{eq:em-e}) can be interpreted as seeking the optimal valid {\it mean parameters} (i.e., expected sufficient statistics) for which the free energy is minimized. 
For discrete latent variables $\y$, the set of all valid mean parameters constitutes a marginal polytope $\mathcal{M}$. In this perspective, the mean-field methods (Equation~\ref{eq:vem-e}) correspond to replacing $\mathcal{M}$ with an {\it inner} approximation $\mathcal{M}'\subseteq\mathcal{M}$. With the restricted set $\mathcal{M}'$ of mean parameters, the E-step generally no longer tightens the bound of the negative marginal log-likelihood, and the algorithm does not necessarily decrease the negative marginal log-likelihood monotonically. However, the algorithm preserves the property that it minimizes the upper bound of the negative marginal log-likelihood. Besides the mean-field methods, there are other approaches for approximation such as belief propagation. These methods correspond to using an {\it outer} approximation $\mathcal{M}''\supseteq\mathcal{M}$ of the marginal polytope, and do not guarantee upper bounds on the negative marginal log-likelihood.

Another approach to restrict the family of $q$ is to assume a parametric distribution $q_{\omega}(\y|\x)$ and optimize the parameters $\bm{\omega}$ in the E-step. The approach has been used in black-box variational inference~\citep{ranganath2014black}, and variational auto-encoders (VAEs)~\citep{kingma2013auto} where $q$ is parameterized as a neural network (a.k.a `inference network,' or `encoder').

It is worth mentioning that the variational approach has also been used for approximate Gaussian processes (GPs, as a nonparametric methods) \citep{titsias2009variational,wilson2016stochastic}, where $\y$ is the inducing points and the variational distribution $q(\y)$ is parameterized as a Gaussian distribution with a nondiagonal covariance matrix that preserves the structures within the true covariance (and hence is different from the above mean-field approximation, which assumes a diagonal variational covariance matrix). We refer interested readers to \citep{wilson2016stochastic} for more details.


{\bf Wake-Sleep.}  
In some cases when the auxiliary $q$ is assumed to have a certain form (e.g., a deep network), the approximate E-step in Equation~\ref{eq:vem-e} may still be too complex to be tractable, or the gradient estimator (w.r.t. the parameters of $q$) can suffer from high variance~\citep{paisley2012variational,mnih2014neural}. To tackle the challenge, more approximations are introduced. 
The wake-sleep algorithm~\citep{hinton1995wake} is one of such methods. In the E-step w.r.t. $q$, rather than minimizing $\KLD(q(\y) \| p_\theta(\y|\x^*))$ (Equation~\ref{eq:em-free-energy}) as in EM and variational EM, the wake-sleep algorithm makes an approximation by minimizing the Kullback–Leibler (KL) divergence in opposite direction: 
\begin{equation}
\begin{split}
    \text{Approximate E-step (Sleep-phase):}\quad \min_{q\in\mathcal{Q}'} \KLD\left( p_\theta(\y|\x^*) \| q(\y) \right),
\end{split}
\label{eq:wake-sleep-e}
\end{equation}
which can be optimized efficiently with gradient descent when $q$ is parameterized.  
Besides wake-sleep, one can also use other methods for low-variance gradient estimation in Equation~\ref{eq:vem-e}, such as reparameterization gradient~\citep{kingma2013auto} and score gradient~\citep{glynn1990likelihood,ranganath2014black,mnih2014neural}. 

In sum, 
the entropy maximization perspective has formulated unsupervised MLE as an optimization-theoretic framework that permits simple alternating minimization solvers. 
Starting from the upper bound of negative marginal log-likelihood (Equation~\ref{eq:em-free-energy}) with maximum entropy and minimum cross entropy, the originally intractable MLE problem gets simplified, and a series of optimization algorithms, ranging from (variational) EM to wake-sleep, arise naturally as an approximation to the original solution.

\subsection{Bayesian Inference}
Now we revisit another classical learning framework, Bayesian inference, and examine its intriguing connections with the maximum entropy principle. Interestingly, the the maximum entropy principle can also help to reformulate Bayesian inference as a constraint optimization problem, as for MLE.  

Different from MLE, Bayesian approach for statistical inference treats the hypotheses (parameters $\btheta$) to be inferred as random variables. Assuming a prior distribution $\pi(\btheta)$ over the parameters, and considering a probabilistic model that defines a conditional distribution $p(\x|\btheta)$, the inference is based on the Bayes' theorem:
\begin{equation}
\begin{split}
    p(\btheta | \mathcal{D}) = \frac{\pi(\btheta)\prod_{\x^*\in\mathcal{D}}p(\x^*| \btheta)}{p(\mathcal{D})},
\end{split}
\label{eq:bayes-rule}
\end{equation}
where $p(\btheta|\mathcal{D})$ is the posterior distribution after observing the data $\mathcal{D}$ (which we assume are i.i.d.); and $p(\mathcal{D}) = \int_{\btheta}\pi(\btheta)\prod_{\x^*} p(\x^*|\btheta) d\btheta$ is the marginal likelihood. 

Interestingly, the early work by \citet{zellner1988optimal} 
showed the relations between Bayesian inference and maximum entropy, by reformulating the statistical inference problem from the perspective of information processing, and rediscovering the Bayes' theorem as the optimal information processing rule. 
More specifically, statistical inference can be seen as a procedure of information processing, where the system receives input information in the form of prior knowledge and data, and emits output information in the form of parameter estimates and others. An efficient inference procedure should generate an output distribution such that the system retains all input information and not inject any extraneous information. The learning objective is thus to minimize the difference between the input and output information w.r.t. the output distribution: 
\begin{equation}
\begin{split}
    \min_{q(\btheta)}&~~ -\ENT(q(\btheta)) + \log p(\dataset) - \E_{q(\btheta)} \left[ \log \pi(\btheta) + \sum\nolimits_{\x^*\in \dataset} \log p(\x^*|\btheta) \right] \\[5pt]
    s.t.&~~ q(\btheta) \in \mathcal{P}(\bm{\Theta}),
\end{split}
\label{eq:bayes-info-proc}
\end{equation}
where the first two terms measure the output information in the output distribution $q(\btheta)$ and marginal $p(\dataset)$, and the third term measures the input information in the prior $\pi(\btheta)$ and data likelihood $p(\x^*|\btheta)$. Here $\mathcal{P}(\bm{\Theta})$ is the space of all probability distributions over $\btheta$.

The optimal solution of $q(\btheta)$ is precisely the the posterior distribution $p(\btheta|\dataset)$ due to the Bayes' theorem (Equation~\ref{eq:bayes-rule}). The proof is straightforward by noticing that the objective can be rewritten as $\min_q \KLD(q(\btheta) \| p(\btheta|\dataset))$.

Similar to the case of duality between MLE and maximum entropy (Equation~\ref{eq:mle-maxent}), the same entropy maximization principle can cast Bayesian inference as a constrained optimization problem. As \citet{jaynes1988comment} commented, this fresh interpretation of Bayes' theorem ``could make the use of Bayesian methods more attractive and widespread, and stimulate new developments in the general theory of inference.'' \citep[][p.280]{jaynes1988comment} The next subsection reviews how entropy maximization as a ``useful tool in generating probability distributions'' \citep[][p.280]{jaynes1988comment} has related to and resulted in more general learning and inference frameworks, such as posterior regularization.

\subsection{Posterior Regularization}\label{sec:background:pr}

The optimization-based formulation of Bayesian inference in Equation~\ref{eq:bayes-info-proc} offers important additional flexibility in learning by allowing rich constraints on machine learning models to be imposed to regularize the outcome. For example, in Equation~\ref{eq:bayes-info-proc} we have seen the standard normality constraint of a probability distribution being imposed on the posterior $q$. It is natural to consider other types of constraints that encode richer problem structures and domain knowledge, which can regularize the model to learn desired behaviors. 

The idea has led to posterior regularization \citep[PR,][]{ganchev2010posterior} or regularized Bayes \citep[RegBayes,][]{zhu2014bayesian}, which augments the Bayesian inference objective with additional constraints:
\begin{equation}
\begin{split}
    \min_{q, \bxi}&~~ - \ENT\left( q(\btheta) \right) - \E_{q(\btheta)} \left[ \sum\nolimits_{\x^*\in \mathcal{D}} \log p(\x^*|\btheta)\pi(\btheta) \right] + U(\bxi) \\[5pt]
    s.t.&~~ q(\btheta) \in \mathcal{Q}(\bxi) \\[3pt]
    &~~\bxi\geq 0,
\end{split}
\label{eq:pr}
\end{equation}
where we have rearranged the terms and dropped any constant factors in Equation~\ref{eq:bayes-info-proc}, and added constraints with $\bxi$ being a vector of slack variables, $U(\bxi)$ a penalty function (e.g., $\ell_1$ norm of $\bxi$), and $\mathcal{Q}(\bxi)$ a subset of valid distributions over $\btheta$ that satisfy the constraints determined by $\bxi$. 
The optimization problem is generally easy to solve when the penalty/constraints are convex and defined w.r.t. a linear operator (e.g., expectation) of the posterior $q$. For example, let $T(\x^*;\btheta)$ be a feature vector of data instance $\x^*\in\dataset$, the constraint posterior set $Q$ can be defined as:
\begin{equation}
\begin{split}
    Q(\bxi) := \left\{ q(\btheta) ~:~ \E_{q}\left[ T(\x^*;\btheta)\right]  \leq \bxi,~ \forall \x^*\in\dataset \right\},
\end{split}
\label{eq:pr-constraint-linear}
\end{equation}
which bounds the feature expectations with $\bxi$. 

Max-margin constraint is another expectation constraint that has shown to be widely effective in classification and regression~\citep{vapnik1998statistical}. The maximum entropy discrimination (MED) by \citet{jaakkola2000maximum} regularizes linear regression models with the max-margin constraints, which is latter generalized to more complex models $p(\x|\btheta)$, such as Markov networks~\citep{taskar2004max} and latent variable models~\citep{zhu2014bayesian}. Formally, let $\y^*\in\mathbb{R}$ be the observed label associated with $\x^*$. The margin-based constraint says that a classification/regression function $h(\x;\btheta)$ should make at most $\epsilon$ deviation from the true label $\y^*$. Specifically, consider the common choice of the function $h$ as a linear function: $h(\x;\btheta) = \btheta^\top T(\x)$, where $T(\x)$ is, with a slight abuse of notation, the feature of instance $\x$. The constraint is written as:
\begin{equation}
\begin{split}
\left\{ \begin{array}{ll}
    \ \ \ \y^* - \E_{q}\left[ \btheta^\top T(\x^*) \right] \leq \epsilon + \xi \\ [4pt]
    -\y^* + \E_{q}\left[ \btheta^\top T(\x^*) \right] \leq \epsilon + \xi',
\end{array}\right.
\end{split}
\end{equation}
for all instances $(\x^*, \y^*)\in\dataset$.


{\bf Alternating optimization for posterior regularization.}
Having seen EM-style alternating minimization algorithms being applied as a general solver for a number of optimization-theoretic frameworks described above, it is not surprising that the posterior regularization framework can also be solved with an alternating minimization procedure. For example, consider the simple case of linear constraint in Equation~\ref{eq:pr-constraint-linear}, penalty function $U(\bxi)=\|\bxi\|_1$, and $q$ factorizing across $\btheta=\{\btheta_c\}$. 
At each iteration $n$, the solution of $q(\btheta_c)$ is given as \citep{ganchev2010posterior}:
\begin{equation}
\begin{aligned}
   \quad q^{(n+1)}(\btheta_c) = \exp\Big\{ \E_{q^{(n)}(\btheta_{\backslash c})} \sum_{\x^*}\log p(\x^*|\btheta)\pi(\btheta) + T(\x^*; \btheta) \Big\} / Z,
\end{aligned}
\label{eq:pr-em-e}
\end{equation}
where $\btheta_{\backslash c}$ denotes all components of $\btheta$ except $\btheta_c$, and $Z$ is the normalization factor. Intuitively, a configuration of $\btheta_{c}$ with a higher expected constraint value $\E_{\backslash c} T(\x^*; \btheta)$ will receive a higher probability under $q^{(n+1)}(\btheta_{c})$. 
The optimization procedure iterates over all components $c$ of $\btheta$.

\subsection{Summary}

In this section, we have seen that the maximum entropy formalism provides an alternative insight into the classical learning frameworks of MLE, Bayesian inference, and posterior regularization. It provides a general expression of these three paradigms as a constrained optimization problem, with a paradigm-specific loss on the model parameters $\btheta$ and an auxiliary distribution $q$, over a properly designed constraint space $\mathcal{Q}$ where $q$ must reside:
\begin{equation}
\begin{split}
    \min_{q, \btheta}&~~ \loss(q, \btheta)
    \\[5pt]
    s.t.&~~ q\in \mathcal{Q}.
\end{split}
\label{eq:background:summary}
\end{equation}
In particular, the use of the auxiliary distribution $q$ converts the originally highly complex problem of directly optimizing $\btheta$ against data, to an alternating optimization problem over $q$ and $\btheta$, which is algorithmically easier to solve since $q$ often acts as an easy-to-optimize proxy to the target model. The auxiliary $q$ can also be more flexibly updated to absorb influence from data or constraints, offering a teacher-student--style iterative  mechanism to incrementally update $\btheta$ as we will see in the sequel.

By reformulating learning as a constrained optimization problem, the maximum entropy point of view 
also offers a great source of flexibility for applying many powerful tools for efficient approximation and enhanced learning, such as variational approximation (e.g., by relaxing $\mathcal{Q}$ to be easy-to-inference family of $q$ such as the mean field family, \citet{jordan1999introduction,xing2002generalized}), convex duality (e.g., facilitating dual sparsity of support vectors via the complementary slackness in the KKT conditions), and kernel methods as used in \citep{taskar2004max,zhu2009maximum}.

It is intriguing that, in the dual point of view on the problem of (supervised) MLE, data instances are encoded as constraints (Equation~\ref{eq:mle-maxent}), much like the structured constraints in posterior regularization. In the following sections, we present the standardized formalism of machine learning algorithms and show that indeed a myriad types of experience besides data instances and constraints can all be encoded in the same generic form and be used in learning.

\section{A Standard Model for Objective Function}\label{sec:se}

Generalizing from Equation~\ref{eq:pr}, we present the following general formulation for learning a target model via a constrained loss minimization program. We would refer to the formulation as the `Standard Equation' because it presents a general space of learning objectives that encompasses many specific formalisms used in different machine learning paradigms. 

Without loss of generality, let $\bt\in\tspace$ be the variable of interest, for example, the input-output pair $\bt=(\x,\y)$ in a prediction task, or the target variable $\bt=\x$ in generative modeling. Let $p_\theta(\bt)$ be the target model with parameters $\btheta$ to be learned. Generally, the SE is agnostic to the specific forms of the target model, meaning that the target model can take an arbitrary form as desired by the problem at hand (e.g., classification, regression, generation, control) and can be of arbitrary types ranging from deep neural networks of arbitrary architectures, prompts for pretrained models, symbolic systems (e.g., knowledge graph), probabilistic graphical models of arbitrary dependence structures, and so on. We discuss more details of the different choices of the target model in Section~\ref{sec:target-model}.

Let $q(\bt)$ be an auxiliary distribution. The SE is written as: 
\begin{equation}
    \begin{split}
        \min_{q,\btheta,\bxi}&~~ - \alpha \BENT\left( q \right) + \beta \BDist\left( q, p_\theta \right)  + U(\bxi) \\[5pt]
        s.t.&~~ - \E_{q}\left[ f^{(\theta)}_k \right] \leq \xi_k,~~ k = 1,\dots,K.
    \end{split}
    \label{eq:se}
\end{equation}
The SE contains three major terms that constitute a learning formalism: the {\it uncertainty function} $\BENT\left( \cdot \right)$ that controls the compactness of the output model (e.g., as measured by the amount of allowed randomness while trying to fit experience); the {\it divergence function} $\BDist\left( \cdot, \cdot \right)$ that measures the distance between the target model to be trained and the auxiliary model that facilitates a teacher--student mechanism as shown below; and the {\it experience function}, which is introduced by a penalty term $U(\bxi)$ that draws in the set of `experience functions' $f_k^{(\theta)}$ that represent external experience of various kinds for training the target model. The hyperparameters $\alpha,\beta\geq 0$ enable trade-offs between these components. 

\begin{figure}
    \centering
    \includegraphics[width=\textwidth]{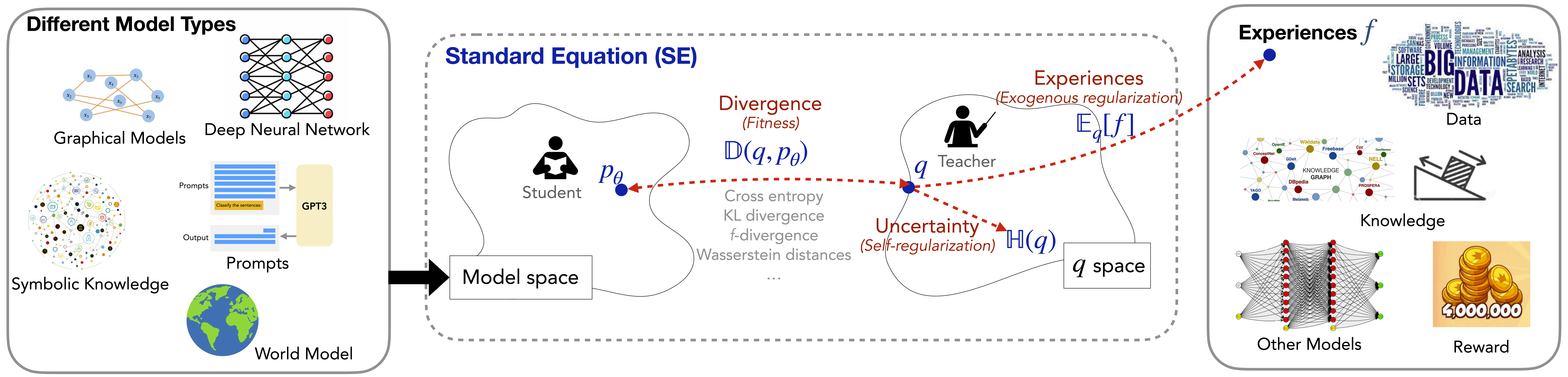}
    \caption{An illustration of the standard equation (SE) as a general formulation of the objective function, used to learn an arbitrary target model $p_\theta$ with any forms of experience. All different forms of experience are formulated uniformly as an experience function $f$. The SE contains three terms, including the experience function for incorporating the exogenous information, the divergence function for improving the fitness of the target model by matching with the auxiliary `teacher' model $q$, and the uncertainty function for controlling the complexity of the learning system.}
    \label{fig:se-overview}
\end{figure}


{\bf Experience function. } 
Perhaps the most powerful in terms of impacting the learning outcome and utility is the experience functions $f_k^{(\theta)}$. An experience function $f^{(\theta)}(\bt)\in \mathbb{R}$ measures the goodness of a configuration $\bt$ in light of any given experience. The superscript ${(\theta)}$ highlights that the experience in some settings (e.g., reward experience as in Section~\ref{sec:experience:reward}) could depend on or be coupled with the target model parameters $\btheta$. In the following, we omit the superscript when there is no ambiguity.
As discussed in Section~\ref{sec:experience}, all diverse forms of experience that can be utilized for model training, such as data examples, constraints, logical rules, rewards, and adversarial discriminators, can be encoded as an experience function. 
The experience function hence provides a unified language to express all exogenous information about the target model, which we consider as an essential ingredient for panoramic learning to flexibly incorporate diverse experience in learning.
Based on the uniform treatment of experience, a standardized optimization program as above can be formulated to identify the desired model. 
Specifically, the experience functions contribute to the optimization objective via the penalty term $U(\bxi)$ over slack variables $\bxi\in\mathbb{R}^K$ applied to the expectation $\E_{q}\left[ f_k \right]$. The effect of maximizing the expectation is such that the auxiliary model $q$ is encouraged to produce samples of 
high quality in light of the experience (i.e., samples receiving high scores as evaluated by the experience function).

{\bf Divergence function. } 
The divergence function $\BDist\left( q, p_\theta \right)$ measures the `quality' of the target model $p_\theta$ in terms of its distance (divergence) with the auxiliary model $q$. Intuitively, we want to minimize the distance from $p_\theta$ to $q$, which is optimized to fit the experience as above. Section~\ref{sec:divergence} gives a concrete example of how the divergence term would directly impact the model training: with a certain specification of the different components (e.g., the experience function, $\alpha/\beta$), the SE in Equation~\ref{eq:se} would reduce to $\min_{\btheta} \BDist(\pdata, p_\theta)$. That is, the learning objective is to minimize the divergence between the target model distribution $p_\theta$ and the data distribution $\pdata$, and the divergence function $\BDist(\cdot, \cdot)$ determines the specific optimization problem. The divergence function can have a variety of choices, ranging from the family of $f$-divergence (e.g., KL divergence), or Bregman divergence, to optimal transport distance (e.g., Wasserstein distance), and so on. We discuss the divergence term in Section~\ref{sec:divergence} in more detail.

{\bf Uncertainty function.} 
The uncertainty function $\BENT(q)$ describes the uncertainty of the auxiliary distribution $q$ and thus controls the complexity of the learning system. It conforms with the maximum entropy principle discussed in Section~\ref{sec:background} that one should pick the most uncertain solution among those that fit all experience. Like other components in SE, the uncertainty measure $\BENT(\cdot)$ can take different forms, such as the popular Shannon entropy, as well as other generalized ones such as Tsallis entropy. In this article, we assume Shannon entropy by default.

\vspace{4pt}

For the discussion in the following sections, it is often convenient to consider a special case of the SE in Equation~\ref{eq:se}. Specifically, we assume a common choice of the penalty $U(\bxi)=\sum_k\xi_k$, and, with a slight abuse of notations, $f=\sum_k f_k$. In this case, the SE in Equation~\ref{eq:se} can equivalently be written in an unconstrained form: 
\begin{equation}
    \begin{split}
        \min_{q,\btheta}&~~ - \alpha \BENT\left( q \right) + \beta \BDist\left( q, p_\theta \right) - \E_{q}\left[ f \right],
    \end{split}
    \label{eq:se-loss}
\end{equation}
which can be easily seen by optimizing Equation~\ref{eq:se} over $\bxi$.
In the special unconstrained form, the interplay between the exogenous experience, divergence, and the endogenous uncertainty become more explicit.

{\bf Optimization: Teacher-student mechanism.}
The introduction of the auxiliary distribution $q$ relaxes the learning problem of $p_\theta$, originally only over $\btheta$, to be now alternating between $q$ and $\btheta$. Here $q$ acts as a conduit between the exogenous experience and the target model: it on the one hand subsumes the experience (by maximizing the expected $f$ value), and on the other hand passes it incrementally to the target model (by minimizing the divergence $\BDist$).
The following fixed point iteration between $q$ and $\btheta$ illustrates this optimization strategy under the SE. Let us plug into Equation~\ref{eq:se-loss} the popular cross entropy (CE) as the divergence function, that is, $\BDist(q,p_\theta)=-\E_q[\log p_\theta]$,  and Shannon entropy as the uncertainty measure, that is, $\BENT(q)=-\E_q[\log q]$.
We further assume the experience $f$ is independent of the model parameters $\btheta$ (the assumption is indeed not necessary for the teacher step). We have, at iteration $n$: 
\begin{equation}
\begin{split}
\text{Teacher:}\quad &q^{(n+1)}(\bt) = \exp\left\{ \frac{\beta \log p_{\theta^{(n)}}(\bt) + f(\bt)}{\alpha} \right\} ~/~ Z \\[5pt]
\text{Student:}\quad &\bm{\theta}^{(n+1)} = \argmax_{\btheta} \E_{q^{(n+1)}(\bt)} \big[ \log p_\theta(\bt) \big],
\end{split}
\label{eq:se-teach-student}
\end{equation}
where $Z$ is the normalization factor. 
The first step embodies a `teacher's update' where the teacher $q$ ingests experience $f$ and builds on current states of the student $p_{\theta^{(n)}}$; the second step is reminiscent of a  `student's update' where the student $p_\theta$ updates its states by maximizing its alignment (here measured by CE) with the teacher. 

Besides, the auxiliary $q$ is an easy-to-manipulate intermediate form in the training that permits rich approximate inference tools for tractable optimization. We have the flexibility of choosing its surrogate functions, ranging from the principled variational approximations for the target distribution in a properly relaxed space (e.g., mean fields) where gaps and bounds can be characterized, to the arbitrary neural network-based `inference networks' that are highly expressive and easy to compute. As can be easily shown (e.g., see Section~\ref{sec:experience:data:unsup}), popular training heuristics, such as EM, variational EM, wake-sleep, forward and backward propagation, and so on, are all direct instantiations or variants of the above teacher-student mechanism with different choices of the form of $q$.

More generally, a broad set of sophisticated algorithms, such as the policy gradient for reinforcement learning and the generative adversarial learning, can also be easily derived by plugging in specific designs of the experience function $f$ and divergence $\BDist$. Table~\ref{tab:instantiation} summarizes various specifications of the SE components that recover a range of existing well-known algorithms from different paradigms.
As shown in more detail in the subsequent sections, the standard equation (Equation~\ref{eq:se} and \ref{eq:se-loss})
offers a unified and universal paradigm for model training under many scenarios based on many types of experience, potentiating a turnkey implementation and a more generalizable theoretical characterization.

\begin{sidewaystable}
\centering
\small
{\renewcommand{\arraystretch}{1.5}
\begin{tabular}{l l l l l l}
\cmidrule[\heavyrulewidth](lr){1-6}
Experience type & Experience function $f$  & Divergence $\mathbb{D}$ & $\alpha$ & $\beta$ &  Algorithm \\ \cmidrule[\heavyrulewidth](lr){1-6}
\multirow{6}{*}{Data instances} & $f_{\text{data}}(\x; \dataset)$ & CE & $1$ & $1$  & Unsupervised MLE \\ 
 & $f_{\text{data}}(\x,\y; \dataset)$ & CE & $1$ & $\epsilon$ & Supervised MLE \\ 
 & $f_{\text{data-self}}(\x,\y; \dataset)$ & CE & $1$ & $\epsilon$ & Self-supervised MLE \\ 
 & $f_{\text{data-w}}(\bt; \dataset)$ & CE & $1$ & $\epsilon$ & Data Re-weighting \\ 
  & $f_{\text{data-aug}}(\bt; \dataset)$ & CE & $1$ & $\epsilon$ & Data Augmentation \\ 
& $f_{\text{active}}(\x,\y; \dataset)$ & CE & $1$ &  $\epsilon$ & Active Learning~\citep{ertekin2007learning}  \\ 
\cmidrule(lr){1-6} 
\multirow{2}{*}{Knowledge} & $f_{rule}(\x,\y)$ & CE & $1$ & $1$  & Posterior Regularization \citep{ganchev2010posterior} \\ 
 & $f_{rule}(\x,\y)$ & CE  & $\mathbb{R}$ & $1$ & Unified EM \citep{samdani2012unified} \\ 
 \cmidrule(lr){1-6} 
\multirow{3}{*}{Reward} & $\log Q^{\theta}(\x,\y)$ & CE  & $1$ & $1$  & Policy Gradient \\ 
 & $\log Q^{\theta}(\x,\y) + Q^{in,\theta}(\x,\y)$ & CE  & $1$ & $1$ & + Intrinsic Reward \\ 
 & $Q^{\theta}(\x,\y)$ & CE & $\rho>0$ & $\rho>0$ & RL as Inference \\ 
 \cmidrule(lr){1-6} 
 Model & $f_{\text{model}}^{\text{mimicking}}(\x,\y; \dataset)$ & CE & $1$ & $\epsilon$ & Knowledge Distillation \citep{hinton2015distilling} \\ 
 \cmidrule(lr){1-6}
\multirow{4}{*}{Variational} & binary classifier & JSD & $0$ & $1$ & Vanilla GAN \citep{goodfellow2014generative} \\ 
& discriminator & $f$-divergence & $0$ & $1$ & f-GAN \citep{nowozin2016f} \\ 
& 1-Lipschitz discriminator & $W_1$ distance & $0$ & $1$ & WGAN \citep{arjovsky2017wasserstein} \\ 
& 1-Lipschitz discriminator & KL & $0$ & $1$ & PPO-GAN \citep{wu2020improving} \\ 
\cmidrule(lr){1-6}
Online & $f_\tau(\bt)$ & CE & $\rho>0$ & $\rho>0$ & Multiplicative Weights \citep{freund1997decision} \\ 
\cmidrule[\heavyrulewidth](lr){1-6}
\end{tabular}
}
\caption{
Example configurations of the components in the standard equation (Eqs.\ref{eq:se}, \ref{eq:se-loss}), which recover different existing algorithms. Here, `CE' means Cross Entropy; `JSD' is the Jensen-Shannon divergence; `$W_1$ dist.' is the first-order Wasserstein distance; and `KL' is the KL divergence.
Refer to Sections~\ref{sec:experience}, \ref{sec:divergence}, and \ref{sec:advanced-experiences} for more details.
}
\label{tab:instantiation}
\end{sidewaystable}

\section{Experience Function}\label{sec:experience}

The experience function $f(\bt)$ in the standard equation can be instantiated to encode vastly distinct types of experience. Different choices of $f(\bt)$ result in learning algorithms applied to different problems. 
With particular choices, the standard equation rediscovers a wide array of well-known algorithms.
The resulting common treatment of the previously disparate algorithms is appealing as it offers new holistic insights into the commonalities and differences of those algorithms. 
Table~\ref{tab:instantiation} shows examples of extant algorithms that are recovered by the standard equation. 

\subsection{Data Instance Experience}\label{sec:experience:data}

\begin{figure}
    \centering
    \includegraphics[width=0.9\textwidth]{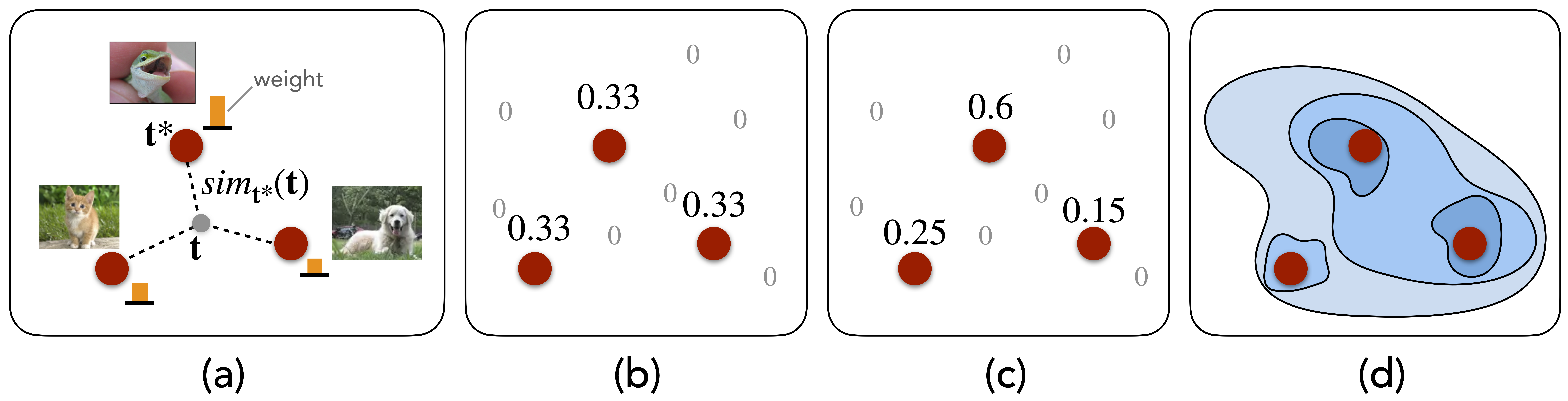}
    \caption{
    Illustration of the data instance experience: {\bf (a)} The general view of the data instance experience function. An observed data set $\dataset=\{\bt^*\}$ (assuming three data instances in the figure) is given. For a configuration $\bt\in\tspace$, its experience function value $f(\bt)$ depends on its similarity with the observed data instances $\bt^*$. Each observed data instance may also be associated with a weight. {\bf (b)} With the indicator function $\indicator_{\bt^*}(\bt)$ as the similarity metric and the default uniform weight of each observed $\bt^*$ (e.g., Equation~\ref{eq:experience:data-sup}), the distribution of the experience function value matches the empirical data distribution (i.e., uniform over the observed instances, and $0$ on all other configurations in $\tspace$). {\bf (c)} By associating different observed instances $\bt^*$ with different weights (Equation~\ref{eq:experience:reweighting}, data reweighting), the experience function corresponds to the reweighted empirical distribution. {\bf (d)} By setting the similarity metric to a soft version (Equation~\ref{eq:experience:augment}, data augmentation), the experience function corresponds to a distribution over $\tspace$ that is more smooth, with nonzero probability on configurations other than the observed data instances.
    }
    \label{fig:experience:data}
\end{figure}

We first consider the most common type of experience, namely, data instances, which are assumed to be independent and identically distributed (i.i.d.). Such data instance experience can appear in a wide range of contexts, including supervised, self-supervised, unsupervised, actively supervised, and other scenarios with data augmentation and manipulation. Figure~\ref{fig:experience:data} illustrates the experience functions based on the data instances.


\subsubsection{\textbf{Supervised data instances}}\label{sec:experience:data:sup}

Without loss of generality and for consistency of notations with the rest of the section, we consider data instances to consist of a pair of input-output variables, namely $\bt=(\x, \y)$. For example, in image classification, $\x$ represents the input image and $\y$ is the object label.
In the supervised setting, we observe the full data drawn i.i.d. from the data distribution $(\x^*,\y^*)\sim\pdata(\x,\y)$. 
For an arbitrary configuration $(\x_0,\y_0)$, its probability $\pdata(\x_0,\y_0)$ under the data distribution can be seen as measuring the \emph{expected similarity} between $(\x_0,\y_0)$ and true data samples $(\x^*,\y^*)$, and be written as $\pdata(\x_0,\y_0)=\E_{\pdata(\x^*,\y^*)}\left[ \indicator_{(\x^*,\y^*)}(\x_0,\y_0) \right]$. Here the similarity measure is $\indicator_{(\x^*,\y^*)}(\x,\y)$, an indicator function that takes the value $1$ if $(\x,\y)$ equals $(\x^*,\y^*)$ and $0$ otherwise (we will see other similarity measures shortly). 
In practice, we are given an empirical distribution $\tilde{p}_{d}(\x,\y)$ by observing a collection of instances $\dataset$ on which the expected similarity is evaluated: 
\begin{equation}
    \begin{split}
        \E_{(\x^*,\y^*) \sim \dataset}\left[ \indicator_{(\x^*,\y^*)}(\x,\y) \right] = \frac{m(\x,\y)}{N},
    \end{split}
    \label{eq:experience:data-empirical}
\end{equation}
where $N$ is the size of the data set $\dataset$, and $m(\x,\y)$ is the number of occurrences of the configuration $(\x,\y)$ in $\dataset$. 

The experience function $f$ accommodates the data instance experience straightforwardly as below:
\begin{equation}
    \begin{split}
        f := f_{\text{data}}(\x,\y; \dataset) = \log \E_{(\x^*,\y^*) \sim \dataset}\left[ \indicator_{(\x^*,\y^*)}(\x,\y) \right].
    \end{split}
    \label{eq:experience:data-sup}
\end{equation}
Figure~\ref{fig:experience:data} (a)-(b) shows an illustration. In particular, the logarithm of the expected similarity is used as the experience function score, that is, the more `similar' a configuration $(\x,\y)$ is to the observed data instances, the higher its quality. The logarithm serves to make the subsequent derivations more convenient as can be seen below.

With this from of $f$, we show that the SE derives the conventional supervised MLE algorithm. 


{\bf Supervised MLE.}
In the SE Equation~\ref{eq:se-loss} (with cross entropy and Shannon entropy), we set $\alpha=1$, and $\beta$ to a very small positive value $\epsilon$. As a result, the auxiliary distribution $q(\x,\y)$ is determined directly by the full data instances (not the model $p_\theta$). That is, the solution of $q$ in the teacher-step (Equation~\ref{eq:se-teach-student}) is:
\begin{equation}
\small
\begin{split}
q(\x,\y) = \exp\left\{ \frac{\beta \log p_{\theta}(\x,\y) + f_{\text{data}}(\x,\y;\dataset)}{\alpha} \right\}/Z \approx \exp\left\{ f_{\text{data}}(\x,\y;\dataset) \right\}/Z = \tilde{p}_{d}(\x,\y),
\end{split}
\label{eq:se-sup-MLE}
\end{equation}
which reduces to the empirical distribution.
The subsequent student-step that maximizes the log-likelihood of samples from $q$ then leads to the supervised MLE updates w.r.t. $\btheta$.


\subsubsection{\textbf{Self-supervised data instances}}\label{sec:experience:data:selfsup}

Given an observed data instance $\bt^* \in \dataset$ in general, one could potentially derive various supervision signals based on the structures of the data and the target model. In particular, one could apply a ``split'' function that artificially partitions $\bt^*$ into two parts $(\x^*,\y^*)=\textit{split}(\bt^*)$ in different, sometimes stochastic ways. Then the two parts are treated as the input and output for the properly designed target model $p_\theta(\x, \y)$ for supervised MLE as above, by plugging in the slightly altered experience function: 
\begin{equation}
    \begin{split}
        f := f_{\text{data-self}}(\x,\y; \dataset) = \log \E_{\bt^*\sim\dataset,\  (\x^*,\y^*) = \textit{split}(\bt^*)}\left[ \indicator_{(\x^*,\y^*)}(\x,\y) \right].
    \end{split}
    \label{eq:experience:data-sup-self}
\end{equation}

A key difference from the above standard supervised learning setting is that now the target variable $\y$ is not costly obtained labels or annotations, but rather part of the massively available data instances. The paradigm of treating part of observed instance as the prediction target is called `self-supervised' learning \citep[e.g.,][]{selfsupervisedblogpost} and has achieved great success in language and vision modeling. For example, in language modeling \citep{devlin2019bert,brown2020language}, the instance $\bt$ is a piece of text, and the `split' function usually selects from $\bt$ one or few words to be the target $\y$ and the remaining words to be $\x$.

\subsubsection{\textbf{Unsupervised data instances}}\label{sec:experience:data:unsup}

In the unsupervised setting, for each instance $\bt=(\x,\y)$, such as (image, cluster index), we only observe the $\x$ part. That is, we are given a data set $\dataset=\{\x^*\}$ without the associated $\y^*$. The data set defines the empirical distribution $\tilde{p}_{d}(\x)$.
The experience can be encoded in the same form as the supervised data (Equation~\ref{eq:experience:data-sup}) but now with only the information of $\x^*$:
\begin{equation}
    \begin{split}
        f := f_{\text{data}}(\x; \dataset) = \log \E_{\x^* \sim \dataset}\left[ \indicator_{\x^*}(\x) \right].
    \end{split}
    \label{eq:experience:data-unsup}
\end{equation}
Applying the SE to this setting with proper specifications derives the unsupervised MLE algorithm.


{\bf Unsupervised MLE.}
The form of Equation~\ref{eq:se-loss} is reminiscent of the variational free energy objective in the standard EM for unsupervised MLE (Equation~\ref{eq:em-free-energy}). We can indeed get exact correspondence by setting $\alpha=\beta=1$, and setting the auxiliary distribution $q(\x,\y)=\tilde{p}_{d}(\x)q(\y|\x)$.
The reason for $\beta=1$, which differs from the specification $\beta=\epsilon$ in the supervised setting, is that the auxiliary distribution $q$ cannot be determined fully by the unsupervised `incomplete' data experience alone. Instead, it additionally relies on $p_\theta$ through the divergence term. 
Here $q$ is assumed a specialized decomposition $q(\x,\y)=\tilde{p}_{d}(\x)q(\y|\x)$ where $\tilde{p}_{d}(\x)$ is fixed and thus not influenced by $p_\theta$. In contrast, if no structure of $q$ is assumed, we could potentially obtain an extended, \emph{instance-weighted} version of EM where each instance $\x^*$ is weighted by the marginal likelihood $p_\theta(\x^*)$, in line with the previous weighted EM methods for robust clustering~\citep[e.g.,][]{gebru2016algorithms,yu2011sample}. 

\subsubsection{\textbf{Manipulated data instances}}\label{sec:experiences:data:manipulate}

Data manipulation, such as reweighting data instances or augmenting an existing data set with new instances, is often a crucial step for efficient learning, such as in a low data regime or in presence of low-quality data sets (e.g., imbalanced labels). 
We show that the rich data manipulation schemes can be treated as experience and be naturally encoded in the experience function~\citep{hu2019learning}. This is done by extending the data-instance experience function (Equation~\ref{eq:experience:data-sup}), in particular by enriching the similarity metric in different ways.  
The discussion here generally applies to data instance $\bt$ of any structures, for example, $\bt=(\x,\y)$ or $\bt=\x$.


{\bf Data reweighting.}
Rather than assuming the same importance of all data instances, we can associate each instance $\bt^*$ with an importance weight $w(\bt^*)\in\mathbb{R}$, so that the learning pays more attention to those high-quality instances, while low-quality ones (e.g., with noisy labels) are downplayed. This can be done by scaling the above 0/1 indicator function (e.g., Equation~\ref{eq:experience:data-sup}) with the weight (Figure~\ref{fig:experience:data}[c]):
\begin{equation}
    \begin{split}
        f := f_{\text{data-w}}(\bt; \dataset) = \log \E_{\bt^* \sim \dataset}\left[ w(\bt^*)\cdot\indicator_{\bt^*}(\bt) \right].
    \end{split}
    \label{eq:experience:reweighting}
\end{equation}
Plugging $f_{\text{data-w}}$ into the SE (Equation~\ref{eq:se-loss}) with the same other specification of supervised MLE ($\alpha=1, \beta=\epsilon$), we get the update rule of model parameters $\btheta$ in the student-step (Equation~\ref{eq:se-teach-student}):
\begin{equation}
    \begin{split}
        \max_{\btheta} \E_{\bt^*\sim\dataset}\left[ w(\bt^*) \cdot \log p_\theta(\bt^*) \right],
    \end{split}
    \label{eq:experience:reweighting-mstep}
\end{equation}
which is the familiar weighted supervised MLE. The weights $w$ can be specified a priori based on heuristics, for example, using inverse class frequency. In many cases it is desirable to automatically induce and adapt the weights during the course of model training. In Section~\ref{sec:app:repurposing}, we discuss how the SE framework can easily enable automated data reweighting by reusing existing algorithms that were designed to solve other seemingly unrelated problems.


{\bf Data augmentation.}
Data augmentation expands existing data by adding synthetically modified copies of existing data instances (e.g., by rotating an existing image at random angles), and is widely used for increasing data size or encouraging invariance in learned representations (e.g., object label is invariant to image rotation). The indicator function $\indicator$ as the similarity metric in Equation~\ref{eq:experience:data-sup} restrictively requires exact match between the true $\bt^*$ and the configuration $\bt$. Data augmentation arises as a `relaxation' to the similarity metric. 
Let $a_{\bt^*}(\bt)\geq 0$ be a distribution that assigns non-zero probability to not only the exact $\bt^*$ but also other configurations $\bt$ related to $\bt^*$ in certain ways (e.g., all rotated images $\bt$ of the observed image $\bt^*$). Replacing the indicator function metric in Equation~\ref{eq:experience:data-sup} with the new $a_{\bt^*}(\bt)\geq 0$ yields the experience function for data augmentation (Figure~\ref{fig:experience:data}[d]):
\begin{equation}
    \begin{split}
        f := f_{\text{data-aug}}(\bt; \dataset) = \log \E_{\bt^* \sim \dataset}\left[ a_{\bt^*}(\bt) \right].
    \end{split}
    \label{eq:experience:augment}
\end{equation}
The resulting student-step updates of $\btheta$, keeping $(\alpha=1, \beta=\epsilon)$ of supervised MLE, is thus:
\begin{equation}
    \begin{split}
        \max_{\btheta} \E_{\bt^* \sim \dataset,~ \bt\sim a_{\bt^*}(\bt)}\left[ \log p_\theta(\bt) \right].
    \end{split}
    \label{eq:experience:augment-mstep}
\end{equation}
The metric $a_{\bt^*}(\bt)$ can be defined in various ways, leading to different augmentation strategies. For example, setting $a_{\bt^*}(\bt)\propto \exp\{ R(\bt,\bt^*) \}$, where $R(\bt,\bt^*)$ is a task-specific evaluation metric such as BLEU for  machine translation, results in the reward-augmented maximum likelihood (RAML) algorithm~\citep{norouzi2016reward}. Besides the manually designed strategies, we can also specify $a_{\bt^*}(\bt)$ as a parameterized transformation process and learn any free parameters thereof automatically (Section~\ref{sec:advanced-experiences}). Notice the same form of the augmentation experience $f_{\text{data-aug}}$ and the reweighting experience $f_{\text{data-w}}$, where the similarity metrics both include learnable components (i.e., $a_{\bt^*}(\bt)$ and $w(\bt^*)$, respectively). Thus the same approach to automated data reweighting can also be applied for automated data augmentation, as discussed more in Section~\ref{sec:app:repurposing}.

\subsubsection{\textbf{Actively supervised data instances}}\label{sec:experience:data:active}
Instead of access to data instances $\x^*$ with readily available labels $\y^*$, in the active supervision setting, we are presented with a large pool of unlabeled instances $\dataset=\{\x^*\}$ as well as a certain budget for querying an oracle (e.g., human annotators) for labeling a limited set of instances. To minimize the need for labeled instances, we need to strategically select queries from the pool according to an {\it informativeness} measure $u(\x)\in\mathbb{R}$. For example, 
$u(\x)$ can be the predictive uncertainty on the instance $\x$, quantified by the Shannon entropy of the predictive distribution or the vote entropy based on a committee of predictors~\citep{dagan1995committee}. 

Mapping the standard equation to this setting, we show the informativeness measure $u(\x)$ is subsumed as part of the experience. Intuitively, $u(\x)$ encodes our heuristic belief about sample `informativeness'. This heuristic is a form of information we inject into the learning system. Denote the oracle as $o$ from which we can draw a label $\y^*\sim o(\x^*)$. The active supervision experience function is then defined as:
\begin{equation}
    \begin{split}
        f := f_{\text{active}}(\x, \y; \dataset) = \log \E_{\x^* \sim \dataset, \y^*\sim o(\x^*)}\left[ \indicator_{(\x^*,\y^*)}(\x, \y) \right] + \lambda \cdot u(\x), 
    \end{split}
    \label{eq:experience:active}
\end{equation}
where the first term is essentially the same as the supervised data experience function (Equation~\ref{eq:experience:data-sup}) with the only difference that now the label $\y^*$ is from the oracle rather than pre-given in $\dataset$; $\lambda>0$ is a trade-off parameter. The formulation of the active supervision is interesting as it is simply a combination of the common supervision experience and the informativeness measure in an {\it additive} manner.

We plug $f_{\text{active}}$ into the SE and obtain the algorithm to carry out learning. The result turns out to recover classical active learning algorithms.


{\bf Active learning.}
Specifically, in Equation~\ref{eq:se-loss}, setting $f=f_{\text{active}}$, and $(\alpha=1,\beta=\epsilon)$ as in supervised MLE, the resulting student-step in Equation~\ref{eq:se-teach-student} for updating $\btheta$ is written as 
\begin{equation}
    \begin{split}
        \max_{\btheta} \E_{\x^*\sim \tilde{p}_{d}(\x)\cdot \exp\{\lambda u(\x)\},~ \y^*\sim o(\x^*)}\left[ \log p_\theta(\x^*,\y^*) \right].
    \end{split}
    \label{eq:experience:active-mstep}
\end{equation}
If the pool $\dataset$ is large, the update can be carried out by the following procedure: we first pick a random subset $\dataset_{\text{sub}}$ from $\dataset$, and select a sample from $\dataset_{\text{sub}}$ according to the informativeness distribution proportional to $\exp\{\lambda u(\x)\}$ over $\dataset_{\text{sub}}$. The sample is then labeled by the oracle, which is finally used to update the target model. By setting $\lambda$ to a very large value (i.e., a near-zero `temperature' $1/\lambda$), we tend to select the {\it most} informative sample from $\dataset_{\text{sub}}$. The procedure rediscovers the algorithm proposed in~\citep{ertekin2007learning} and more generally the pooling-based active learning algorithms~\citep{settles2012active}.

\subsection{Knowledge-Based Experience}\label{sec:experience:knowledge}

Many aspects of problem structures and human knowledge are difficult if not impossible to be expressed through individual data instances. Examples include the knowledge of expected feature values, maximum margin structures (Section~\ref{sec:background:pr}), logical rules, and so on. The knowledge generally imposes constraints that we want the target model to satisfy. The experience function in the standard equation is a natural vehicle for incorporating such knowledge constraints in learning. Given a configuration $\bt$, the experience function $f(\bt)$ measures the degree to which the configuration satisfies the constraints. 

As an example, we consider 
first-order logic (FOL) rules, which provide an expressive declarative language to encode complex symbolic knowledge~\citep{hu2016harnessing}. More concretely, let $f_{\text{rule}}(\bt)$ be an FOL rule w.r.t. the variables $\bt$. For flexibility, we use soft logic~\citep{bach2017hinge} to formulate the rule. Soft logic allows continuous truth values from the interval $[0,1]$ instead of $\{0, 1\}$, and the Boolean logical operators are redefined as: 
\begin{equation}
\begin{split}
A \& B = \max\{A+B-1, 0\}&,\quad A\vee B =\min\{A+B,1\} \\
A_1 \wedge \dots \wedge A_N = \sum\nolimits_{i}A_i / N&,\quad \neg A = 1 - A.
\end{split}
\label{eq:experience:knowledge:soft-logic}
\end{equation}
Here $\&$ and $\wedge$ are two different approximations to logical conjunction: $\&$ is useful as a selection operator (e.g., $A\&B=B$ when $A=1$, and $A\&B=0$ when $A=0$), while $\wedge$ is an averaging operator. To give a concrete example, consider the problem of sentiment classification, where given a sentence $\x$, we want to predict its sentiment $\y \in \{\text{negative}\ 0,\text{positive}\ 1\}$. A challenge for a sentiment classifier is to understand the contrastive sense within a sentence and capture the dominant sentiment precisely. For example, if a sentence is of structure `A-but-B' with the connective `but', the sentiment of the half sentence after `but' dominates. Let $\x_B$ be the half sentence after `but' and $\tilde{\y}_B\in[0,1]$ the (soft) sentiment prediction over $\x_B$ by the current model, a possible way to express the knowledge as a logical rule $f_{\text{rule}}(\x, \y)$ is:
\begin{equation}
\begin{split}
f := f_{\text{rule}}(\x,\y) = \text{has-`A-but-B'-structure}(\x) \Rightarrow (\indicator(\y=1) \Rightarrow \tilde{\y}_B ~~\&~~ \tilde{\y}_B \Rightarrow \indicator(\y=1)),
\end{split}
\label{eq:experience:knowledge:sentiment}
\end{equation}
where $\indicator(\cdot)$ is an indicator function that takes $1$ when its argument is true, and 0 otherwise. Given an instantiation (a.k.a. grounding) of $(\x, \y, \tilde{\y}_B)$, the truth value of $f_{\text{rule}}(\x, \y)$ can be evaluated by definitions in Equation~\ref{eq:experience:knowledge:soft-logic}. Intuitively, the $f_{\text{rule}}(\x, \y)$ truth value gets closer to $1$ when $\y$ and $\tilde{\y}_B$ are more consistent.

We then make use of the knowledge-based experience such as $f_{\text{rule}}(\bt)$ to drive learning. The standard equation rediscovers classical algorithms for learning with symbolic knowledge. 


{\bf Posterior regularization and extensions.}
By setting $\alpha=\beta=1$ and $f$ to a constraint function such as $f_{\text{rule}}$, the SE with cross entropy naturally leads to a generalized posterior regularization framework~\citep{hu2016harnessing}: 
\begin{equation}
\begin{split}
    \min_{\btheta, q}&~~ - \ENT\left( q(\bt) \right) - \E_{q(\bt)} \left[ \log p_\theta(\bt) \right] - \E_{q(\bt)}\left[ f_{\text{rule}}(\bt)  \right],
\end{split}
\label{eq:experience:knowledge:pr}
\end{equation}
which extends the conventional Bayesian inference formulation (Section~\ref{sec:background:pr}) by permitting regularization on arbitrary random variables of arbitrary models (e.g., deep neural networks) with complex rule constraints.

The trade-off hyperparameters can also take other values. For example, by allowing arbitrary $\alpha\in\mathbb{R}$, the objective corresponds to the {\it unified expectation maximization} (UEM) algorithm~\citep{samdani2012unified} that extends the posterior regularization 
for added flexibility.

\subsection{Reward Experience}\label{sec:experience:reward}

\begin{wrapfigure}{r}{0.5\textwidth}
    \centering
    \includegraphics[width=0.47\textwidth]{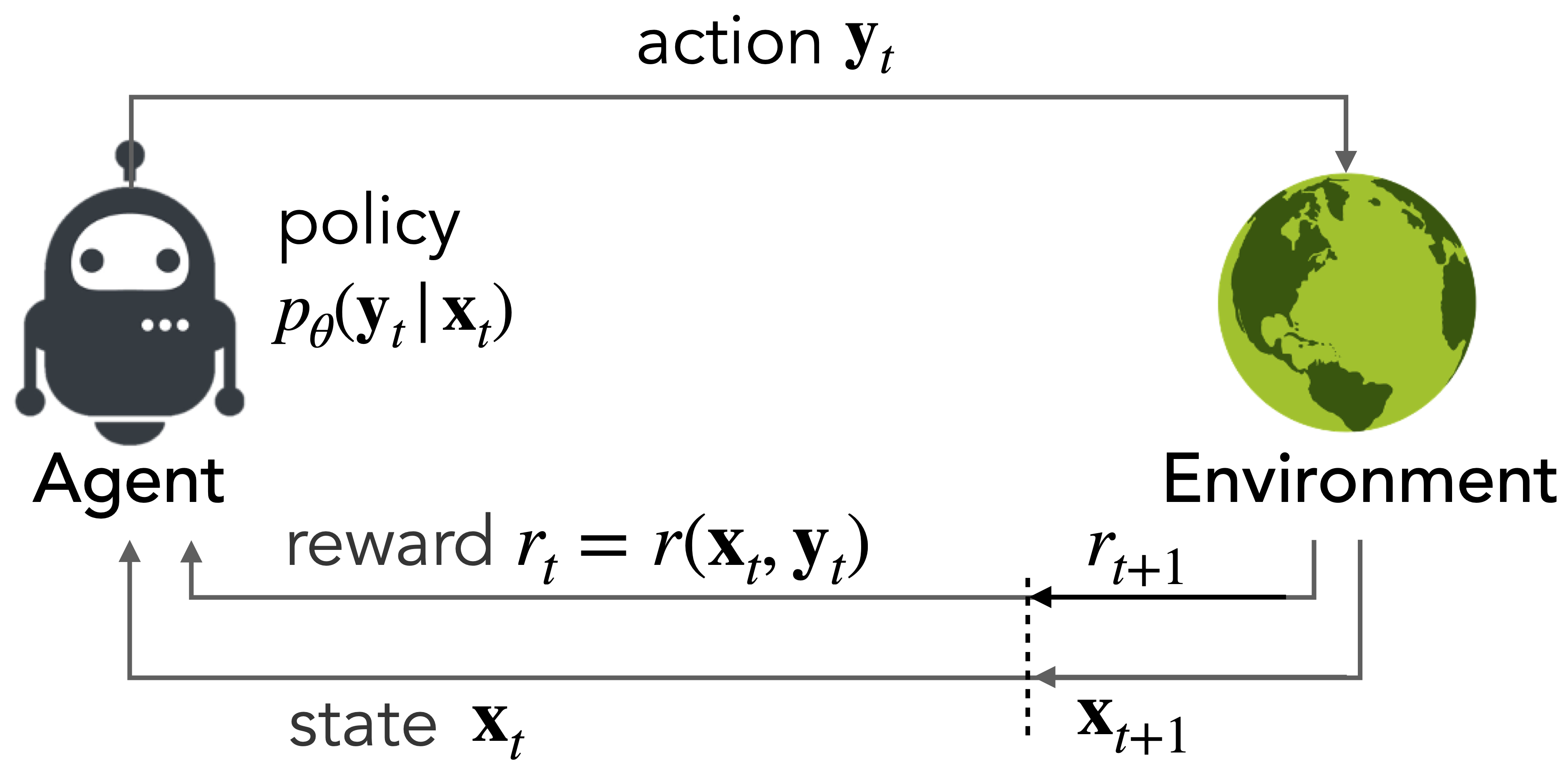}
    \caption{The MDP setting.}
    \label{fig:mdp}    
\end{wrapfigure}

We now consider a very different learning setting commonly seen in robotic control and other sequential decision making problems. In this setting, experience is gained by the agent interacting with external environment and collecting feedback in the form of rewards.
Formally, we consider a Markov decision process (MDP) as illustrated in Figure~\ref{fig:mdp}, where $\bt=(\x, \y)$ is the state-action pair. For example, in playing a video game, the state $\x$ is the game screen by the environment (the game engine) and $\y$ can be any game actions.
At time $t$, the 
environment is in 
state $\x_t$. The agent draws an action $\y_t$ according to the policy $p_\theta(\y|\x)$. The state subsequently transitions to $\x_{t+1}$ following certain transition dynamics of the environment, and yields a reward $r_t=r(\x_t, \y_t)\in \mathbb{R}$. 
The general goal of the agent is to learn the policy $p_\theta(\y|\x)$ to maximize the reward in the long run. There could be different specifications of the goal.
In this section we focus on the one where we want to maximize the expected discounted reward starting from a state drawn from an arbitrary state distribution $p_0(\x)$, with a discount factor $\gamma\in[0,1]$ applied to future rewards. 

A base concept that plays a central role in characterizing the learning in this setting is the {\it action value function}, also known as the $Q$ function, which is the expected discounted future reward of taking action $\y$ in state $\x$ and continuing with the policy $p_{\theta}$:
\begin{equation}
    \begin{split}
        Q^{\theta}(\x,\y)= \E\left[ \sum_{t=0}^\infty\gamma^t r_t ~|~ \x_0=\x,\y_0=\y \right], 
    \end{split}
    \label{eq:experience:reward}
\end{equation}
where the expectation is taken by following the state dynamics induced by the policy (thus the dependence of $Q^{\theta}$ on policy parameters $\btheta$). 
We next discuss how $Q^{\theta}(\x,\y)$ can be used to specify the experience function in different ways, which in turn derives various known algorithms in reinforcement learning (RL) \citep{sutton2017reinforcement}.
Note that here we are primarily interested in learning the conditional model (policy) $p_\theta(\y|\x)$. Yet we can still define the joint distribution as $p_\theta(\x,\y)=p_\theta(\y|\x)p_0(\x)$.

\subsubsection{\textbf{Expected future reward}}\label{sec:experience:reward:Q}
The first simple way to use the reward signals as the experience is by defining the experience function as the logarithm of the expected future reward:
\begin{equation}
    \begin{split}
        f := f^{\theta}_{\text{reward},1}(\x, \y) = \log Q^{\theta}(\x,\y),
    \end{split}
    \label{eq:experience:reward-pg}
\end{equation}
which leads to the classical policy gradient algorithm~\citep{sutton2000policy}.


{\bf Policy gradient.}
With $\alpha=\beta=1$, we arrive at policy gradient. To see this, consider the teacher-student optimization procedure in Equation~\ref{eq:se-teach-student}, where the teacher-step yields the $q$ solution: 
\begin{equation}
    \begin{split}
    q^{(n)}(\x,
\y) = p_{\theta^{(n)}}(\x, \y) Q^{\theta^{(n)}}(\x,\y) ~/~ Z,
     \end{split}
    \label{eq:experience:reward-pg-estep}
\end{equation}   
and the student-step updates $\btheta$ with the gradient at $\btheta=\btheta^{(n)}$:
\begin{equation}
    \begin{split}
        &\E_{q^{(n)}(\x,\y)}\left[ \nabla_\theta \log p_{\theta}(\x,\y) \right] + \E_{q^{(n)}(\x,\y)}\left[ \nabla_\theta f^{\theta}_{\text{reward},1}(\x,\y) \right] ~\Big|\Big._{\btheta=\btheta^{(n)}} \\
        &= 1/Z \cdot \sum_{\x} p_0(\x) \nabla_\theta \sum_{\y} p_\theta(\y|\x) Q^{\theta}(\x,\y) ~\Big|\Big._{\btheta=\btheta^{(n)}} \\
        &= 1/Z \cdot \sum_{\x} \mu^{\theta}(\x) \sum_{\y} Q^{\theta}(\x,\y) \nabla_\theta p_\theta(\y|\x) ~\Big|\Big._{\btheta=\btheta^{(n)}}.
    \end{split}
    \label{eq:experience:reward-pg-mstep}
\end{equation}
Here the first equation is due to the log-derivative trick $g \nabla \log g = \nabla g$; and the second equation is due to the policy gradient theorem~\citep{sutton2000policy}, where $\mu^{\theta}(\x)=\sum_{t=0}^{\infty} \gamma^t p(\x_t=\x)$ is the unnormalized discounted state visitation measure. 
The final form is exactly the policy gradient up to a multiplication factor $1/Z$.

We can also consider a slightly different use of the reward, by directly setting the experience function to the $Q$ function:
\begin{equation}
    \begin{split}
        f := f^{\theta}_{\text{reward},2}(\x, \y) = Q^{\theta}(\x,\y).
    \end{split}
    \label{eq:experience:reward-inference}
\end{equation}
This turns out to connect to the known RL-as-inference approach that has a long history of research~\citep[e.g.,][]{dayan1997using,deisenroth2013survey,rawlik2012stochastic,levine2018reinforcement,abdolmaleki2018maximum}.

{\bf RL as inference.}
We set $\alpha=\beta:=\rho>0$.
The configuration corresponds to the approach that casts RL as a probabilistic inference problem. To see this, we introduce an additional binary random variable $o$, with $p(o=1|\x,\y)\propto\exp\{Q(\x,\y) / \rho\}$. Here $o=1$ is interpreted as the event that maximum reward is obtained, $p(o=1|\x,\y)$ is seen as the `conditional likelihood', and $\rho$ is the temperature. The goal of learning is to maximize the marginal likelihood of optimality: $\log p(o=1)$, which, however, is intractable to solve. Much like how the standard equation applied to unsupervised MLE provides a surrogate variational objective for the marginal data likelihood (Section~\ref{sec:experience:data:unsup}), here the standard equation also derives a variational bound for $\log p(o=1)$ (up to a constant factor) with the above specification of $(f,\alpha,\beta)$: 
\begin{equation}
    \begin{split}
        - \log p(o=1) 
&= - \log \E_{p_\theta(\x,\y)}\left[ p(o=1|\x,\y) \right] \\
&\leq - \rho \ENT\left( q \right) - \rho \E_{q(\x,\y)}\left[ \log p_\theta(\x,\y) \right] - \E_{q(\x,\y)}\left[ Q^{\theta}(\x,\y) \right].
    \end{split}
    \label{eq:experience:reward-inference-elbo}
\end{equation}
Following the teacher-student procedure in Equation~\ref{eq:se-teach-student}, the teacher-step produces the $q$ solution:
\begin{equation}
    \begin{split}
    q^{(n)}(\x,
\y) = p_{\theta^{(n)}}(\x, \y) \exp \left\{ Q^{\theta^{(n)}}(\x,\y) / \rho \right\} ~/~ Z.
     \end{split}
    \label{eq:experience:reward-inference-estep}
\end{equation}   
The subsequent student-step involves approximation by fixing $\btheta=\btheta^{(n)}$ in $Q^{\theta}(\x,\y)$ in the above variational objective, and minimizes only $\E_{q^{(n)}(\x,\y)}\left[ \log p_\theta(\x,\y) \right]$ w.r.t. $\btheta$.

\subsubsection{\textbf{Intrinsic reward}}\label{sec:experience:reward:intrinsic}

Rewards provided by the extrinsic environment can be sparse in many real-world sequential decision problems. Learning in such problems is thus difficult due to the lack of supervision signals. A method to alleviate the difficulty is to supplement the extrinsic reward with dense {\it intrinsic} reward that is generated by the agent itself (i.e., the agent is intrinsically motivated). The intrinsic reward can be induced in various ways, such as the `curiosity'-based reward that encourages the agent to explore novel or `surprising' states~\citep{schmidhuber2010formal,houthooft2016vime,pathak2017curiosity}, or the `optimal reward', which is designed with the goal of encouraging maximum extrinsic reward at the end~\citep{singh2010intrinsically,zheng2018learning}. Formally, let $r_t^{in}=r^{in}(\x_t, \y_t)\in\mathbb{R}$ be the intrinsic reward at time $t$ with state $\x_t$ and action $\y_t$. For example, in \citep{pathak2017curiosity}, $r^{in}_t$ is the prediction error (i.e., the `surprise') of the next state $\x_{t+1}$. Let $Q^{in,\theta}(\x,\y)$ denote the action-value function for the intrinsic reward, defined in a similar way as the extrinsic $Q^{\theta}(\x,\y)$:
\begin{equation}
    \begin{split}
        Q^{in,\theta}(\x,\y) = \E\left[ \sum_{t=0}^\infty\gamma^t r_t^{in} ~|~ \x_0=\x,\y_0=\y \right].
    \end{split}
    \label{eq:experience:reward-intrinsic}
\end{equation}
It is straightforward to derive the intrinsically motivated variant of the policy gradient algorithm (and other RL algorithms discussed below), by replacing the standard extrinsic-only $Q^{\theta}(\x,\y)$ in the experience function Equation~\ref{eq:experience:reward-pg} with the combined $Q^{\theta}(\x,\y) + Q^{in,\theta}(\x,\y)$. 
Let $f^{\theta}_\text{reward,ex+in}(\x,\y)$ denote the resulting experience function that incorporates both the extrinsic and the additive intrinsic rewards.

We can notice some sort of symmetry between $f^{\theta}_\text{reward,ex+in}(\x,\y)$ and the actively supervised data experience $f_\text{active}$ in Equation~\ref{eq:experience:active}, which augments the standard supervised data experience with the additive informativeness measure $u(\x)$. 
The resemblance could naturally inspire mutual exchange between the research areas of intrinsic reward and active learning, for example, using the active learning informativeness measure as the intrinsic reward  $r^{in}$, as was studied in earlier work~\citep{schmidhuber2010formal,li2011knows,pathak2019self}.

\subsection{Model-Based Experience}\label{sec:experinece:model}

A model may also learn from other models of the same or related tasks. For example, one can learn a small-size target model by mimicking the outputs of a larger pretrained model, or an ensemble of multiple models, that is more powerful but often too expensive to deploy. Thus, the large model serves as the experience, or the source of information about the task at hand. By seeing that the large source model is effectively providing `pseudo-labels' on the observed inputs $\dataset=\{\x^*\}$, we can readily write down the corresponding experience function, as a variant of the standard supervised data experience function in Equation~\ref{eq:experience:data-sup}:
\begin{equation}
\begin{split}
f := f_{\text{model}}^{\text{mimicking}}(\x, \y; \dataset) = \log \E_{\x^* \sim \dataset, \tilde{\y}\sim p_{\text{model}'}(\y|\x^*)}\left[ \indicator_{(\x^*,\tilde{\y})}(\x,\y)  \right],
\end{split}
\label{eq:experience:model-mimic}
\end{equation}
where $\tilde{\y}\sim p_{\text{model}'}(\y|\x^*)$ denotes that we draw label samples from the output distribution of the large source model.

Another way of model-based experience is that the pretrained model directly measures the score of a given configuration $(\x,\y)$ with its (log-)likelihood:
\begin{equation}
\begin{split}
f := f_{\text{model}}^{\text{scoring}}(\x, \y) = \log p_{\text{model}'}(\y|\x).
\end{split}
\label{eq:experience:model-score}
\end{equation}
As a concrete example, consider learning a text generation model that aims to generate text $\y$ conditioning on a given sentiment label $\x$ (either positive or negative). A natural form of experience that has encoded the concept of `sentiment' is a pretrained sentiment classifier, which can be used to measure the likelihood (plausibility) that the given sentence $\y$ has the given sentiment $\x$ \citep{hu2017controllable}. The likelihood serves as the experience function score used to drive target model training through the SE.

{\bf Knowledge distillation.}
Plugging the model-based experience function $f_{\text{model}}^{\text{mimicking}}$ (Equation~\ref{eq:experience:model-mimic}) into SE rediscovers the well-known knowledge distillation algorithm \citep{hinton2015distilling}. Specifically, by following the same SE specification of the supervised MLE (Section~\ref{sec:experience:data:sup}) and setting $f=f_{\text{model}}^{\text{mimicking}}$, we obtain the knowledge distillation objective:
\begin{equation}
\begin{split}
\E_{\x^*\sim\dataset, \tilde{\y}\sim p_{\text{model}'}(\y|\x^*)} \left[ \log p_\theta(\tilde{\y} | \x^*) \right],
\end{split}
\label{eq:experience:model:knowledge-distill}
\end{equation}
which trains the target model $p_\theta$ by encouraging it to mimic the source model outputs (and thus the source model is also called `teacher' model---in a similar sense to but not to be confused with the teacher-student mechanism for optimization described in Sections~\ref{sec:se} and \ref{sec:opt}).

\section{Divergence Function}\label{sec:divergence}

We now turn to the divergence term $\BDist(q,p_\theta)$ that measures the distance between the auxiliary distribution $q$ and the model distribution $p_\theta$ in the SE. The discussion in the prior section has assumed specific case of $\BDist$ being the cross entropy. Yet there is a rather rich set of choices for the divergence function, such as $f$-divergence (e.g., KL divergence, Jensen-Shannon divergence), optimal transport distance (e.g., Wasserstein distance), and so on.

To see a concrete example of how the divergence function may influence the learning, consider the experience to be data instances with the data distribution $\pdata(\bt)$, and, following the configurations of supervised MLE (Section~\ref{sec:experience:data:sup}), set $f=f_{\text{data}}$, $\alpha=1$, $\beta=\epsilon$, and the uncertainty measure $\BENT$ to be the Shannon entropy. As a result, the solution of $q$ in Equation~\ref{eq:se-loss} reduces to the data distribution $q(\bt)=\pdata(\bt)$. The learning of the model thus reduces to minimizing the divergence between the model and data distributions:
\begin{equation}
\begin{split}
    \min_{\btheta} \BDist(\pdata, p_\theta),
\end{split}
\label{eq:divg-objective}
\end{equation}
which is a common objective shared by many ML algorithms depending on how the divergence function is specialized. Thus, in this setting, the divergence function directly determines the learning objective. In the following sections, we will see other richer influences of $\BDist$ in combination with other SE components.

We next discuss some of the common choices for the divergence function, which opens up the door to recover and generalize more learning algorithms besides those discussed earlier, such as the generative adversarial learning \citep[e.g., GANs,][]{goodfellow2014generative} that is widely used to simulate complex distributions (e.g., natural image distributions).

\subsection{Cross Entropy and Kullback–Leibler (KL) Divergence}\label{sec:divergence:ce}
The diverse algorithms discussed in Section~\ref{sec:experience} have all been based on the cross entropy as the divergence function in SE, namely,
\begin{equation}
\begin{split}
    \BDist(q, p_\theta) = - \E_q \left[ \log p_\theta\right].
\end{split}
\label{eq:divg:ce}
\end{equation}
A nice advantage of using the cross entropy is the close-form solution of $q$ in the teacher-student procedure, as shown in Equation~\ref{eq:se-teach-student}, which makes the optimization and analysis easier. 

In the case where the uncertainty function is the Shannon entropy $\BENT(q)=-\E_q[\log q]$ (as is commonly assumed in this article), one could alternatively see the above algorithms as using the KL divergence for $\BDist$, by noticing that $\KLD(q, p_\theta) = \E_{q}[ \log q] - \E_{q}[\log p_\theta]$. That is, given specific balancing weights $(\alpha_0, \beta_0)$, the divergence and uncertainty terms in SE can be rearranged as:
\begin{equation}
\begin{split}
    \alpha_0 \E_q[\log q] - \beta_0 \E_q \left[ \log p_\theta\right] := (\alpha_0 - \beta_0) \E_q \left[ \log p_\theta\right] + \beta_0 \KLD\left( q, p_\theta\right),
\end{split}
\label{eq:divg:kl}
\end{equation}
where the KL divergence term corresponds to $\BDist$ in SE if we see $\alpha = \alpha_0 - \beta_0$ and $\beta = \beta_0$.

\subsection{Jensen-Shannon (JS) Divergence}\label{sec:divergence:jsd}\label{sec:divergence:gans}
JS divergence provides another common choice for the divergence function:
\begin{equation}
\begin{split}
    \BDist(q, p_\theta) = \text{JS}\left( q \| p_\theta \right) = \frac{1}{2} \KLD\left( q \| h \right) + \frac{1}{2} \KLD\left( p_\theta \| h \right),
\end{split}
\label{eq:divg:jsd}
\end{equation}
where $h := \frac{1}{2} (q + p_\theta)$ is the mean distribution. In particular, by considering the specific instantiation in Equation~\ref{eq:divg-objective} of the SE and setting $\BDist$ to the JS divergence, we can derive the algorithm for learning the generative adversarial networks \citep[GANs,][]{goodfellow2014generative} as shown below. From this perspective, the key concept in generative adversarial learning, namely the {\it discriminator}, arises as an approximation to the optimization procedure. We discuss in Section~\ref{sec:advanced-experiences} an alternative view of the learning paradigm where the discriminator plays the role of `dynamic' experience in SE.

{\bf Generative adversarial learning: The functional descent view.}
To optimize the objective in Equation~\ref{eq:divg-objective} with the JS divergence, {\it probability functional descent} \citep[PFD,][]{chu2019probability} offers an elegant way that recovers the optimization procedure of GANs originally developed in \citet{goodfellow2014generative}. Here we give the PFD result directly, and provide a more detailed review of the PFD optimization in Section~\ref{sec:opt}.

Specifically, let $J(p):=\BDist(\pdata, p)$ in Equation~\ref{eq:divg-objective}, which is a functional on the distribution $p \in \mathcal{P}(\tspace)$. The PFD approach \citep{chu2019probability} shows that minimizing $J(p)$ w.r.t $p$ can equivalently be done by solving the following saddle-point problem:
\begin{equation}
\begin{split}
    \inf_{p} \sup_{\varphi} \E_{p}\left[ \varphi(\bt) \right] - J^*(\varphi),
\end{split}
\label{eq:divg:pfd}
\end{equation}
where $\varphi: \tspace\to\mathbb{R}$ is a continuous function, and $J^*(\varphi) = \sup\nolimits_{h} \E_{h} \left[ \varphi(\bt) \right] - J(h)$ is the convex conjugate of $J$. In the case of $J(p)$ being the JS divergence with $\pdata$, if we approximate the above optimization by parameterizing $\varphi(\bt)$ as $\varphi_{\phi}=\frac{1}{2}\log(1-C_\phi)-\frac{1}{2}\log 2$ where $C_\phi: \tspace\to[0,1]$ is a binary classifier (a.k.a., discriminator) and $p$ as $p_\theta$ (i.e., the target model), Equation~\ref{eq:divg:pfd} recovers the original GAN algorithm~\citep{goodfellow2014generative}:
\begin{equation}
\begin{split}
    \min_{\btheta} \max_{\bphi} \frac{1}{2} \E_{\pdata}\left[ \log C_\phi(\bt) \right] - \frac{1}{2} \E_{p_\theta} \left[ \log ( 1 - C_\phi(\bt) ) \right].
\end{split}
\label{eq:divg:pfd-gan}
\end{equation}

\subsection{Wasserstein Distance}\label{sec:divergence:wasserstein}

Another distance measure that is receiving increasingly interest is the Wasserstein distance, a member of the optimal transport distance family \citep{peyre2019computational,santambrogio2015optimal}. Compared to many of the divergence metrics (e.g., KL divergence),  Wasserstein distance has the desirable properties as a distance metric, such as symmetry and the triangle inequality. Based on the Kantorovich duality \citep[][Section 1.2]{santambrogio2015optimal}, the first-order Wasserstein distance between the two distributions $q$ and $p$ can be written as:
\begin{equation}
\begin{split}  
    W_1(q, p) = \sup_{\|\varphi\|_L \leq 1} \E_{q}\left[ \varphi(\bt) \right] - \E_{p}\left[ \varphi(\bt) \right],
\end{split}
\label{eq:divg:w-dist}
\end{equation}
where $\|\varphi\|_L\leq 1$ is the constraint of $\varphi: \tspace\to \mathbb{R}$ being a 1-Lipschitz function. 

{\bf Wasserstein GAN.}
Setting the divergence function in Equation~\ref{eq:divg-objective} to the Wasserstein distance $W_1$, we thus recover the Wasserstein GAN algorithm \citep{arjovsky2017wasserstein}:
\begin{equation}
\begin{split}
    \min_{\btheta} W_1(p_d, p_\theta) = \min_{\btheta} \sup_{\|\varphi\|_L \leq 1} \E_{p_d}\left[ \varphi(\bt) \right] - \E_{p_\theta}\left[ \varphi(\bt) \right],
\end{split}
\label{eq:divg:wgan}
\end{equation}
which is shown to be more robust than the original GAN algorithm based on the JS divergence (Section~\ref{sec:divergence:jsd}).

\vspace{20pt} 

\section{Dynamic SE}\label{sec:advanced-experiences}

The standard equation Equation~\ref{eq:se} so far has played the role of the ultimate learning objective that fully defines the learning problem in an analytical form. As seen in Sections~\ref{sec:experience} and \ref{sec:divergence}, many of the known algorithms are special cases of the SE objective. On the other hand, in a dynamic or online setting, the learning objective itself may be evolving over time. For example, the data instances may follow changing distributions or come from evolving tasks (e.g., a sequence of tasks that are increasingly complex); the experience in a strategic game context can involve complex interactions with the target model through co-training or adversarial dynamics; and the success criteria of the model w.r.t. the experience can be adapting. In this section, we discuss an {\it extended} view of the SE in dealing with the learning in such dynamic contexts. Instead of serving as the static overall objective function, now the SE is a core part of an outer loop in the learning procedure.

More specifically, each of the SE components (e.g., experience function, divergence function, balancing weights) can change over time. For example, consider a dynamic experience function $f_\tau$, which is indexed by $\tau$ indicating its evolution over time, iterations, or tasks. This differs from the experience discussed earlier in Section~\ref{sec:experience} that is defined a priori (e.g., a static set of data instances) and encoded as a fixed experience function $f$. With the dynamic experience, 
a learning procedure, using the special case of SE in Equation~\ref{eq:se-loss}, can be written as:
\begin{equation}
\begin{split}
\text{for}~ &\tau = 1, 2, ...: \\
    &\text{Acquire experience $f_\tau$}, \\
    &\text{Solve SE:}~~ \min_{q,\btheta} - \alpha \BENT\left( q \right) + \beta \BDist\left( q, p_\theta \right) - \E_{q}\left[ f_\tau \right].
\end{split}
\label{eq:dynamic-se}
\end{equation}
Here the SE governs the optimization of the target model $p_\theta$ given the updated experience at each $\tau$. We discuss in more detail the SE with dynamic experience (Section~\ref{sec:dynamic:experiences}) and other dynamic components (Section~\ref{sec:dynamic:other}), which further recovers several well-known algorithms in existing learning paradigms.

\subsection{Dynamic Experience}\label{sec:divergence:gans:var}\label{sec:dynamic:experiences}

The experience can change over time due to different reasons. For example, the experience can involve optimization, often together with the target model, resulting in a bi-level optimization scheme (Section~\ref{sec:dynamic:experience:variational}); or alternatively, the experience may just come from the environment sequentially with an unknown dynamic (Section~\ref{sec:dynamic:experience:online}).

\subsubsection{\textbf{Variational Experience With Optimization}}\label{sec:dynamic:experience:variational}

As a concrete example, consider the experiences that do not have an analytic form, but instead are defined in a variational way (i.e., as a solution to an optimization problem):
\begin{equation}
\begin{split}
    f_\tau := \argmax_f \mathcal{J}(f, p_{\theta^{(\tau-1)}}),
\end{split}
\label{eq:experience:advanced}
\end{equation}
with an optimization objective $\mathcal{J}$, where $p_{\theta^{(\tau-1)}}$ is the target model learned with the experience from the last iteration.

A particular example is the \emph{adversarial} experience emergingly used in many generation and representation learning problems. Specifically, recall the 
data instance experience $f_{\text{data}}(\bt;\dataset)$ that measures the closeness between a configuration $\bt$ with the true data $\dataset$ based on data instance matching (Equation~\ref{eq:experience:data-sup}). Such manually defined measures could be subjective, suboptimal, or demanding expertise or heavy engineering to be properly specified. 
An alternative way that sidesteps the drawbacks is to automatically induce a closeness measure $f_\phi(\bt)$, where $\bphi$ denotes any free parameters associated with the experience and is to be learned. For example, one can measure the closeness of a configuration $\bt$ to the data set $\dataset$ based on a 
\emph{discriminator} (or \emph{critic}) that evaluates how easily $\bt$ can be differentiated from the instances in $\dataset$.
A concrete application is in image generation, where a binary discriminator takes as input an image sample and tells whether the input image is a real instance from the observed image corpus $\dataset$ or a fake one produced by the model. The similar idea of discriminator-based closeness measure was also explored in the likelihood-free inference literature~\citep{gutmann2018likelihood}. 

The discriminator/critic as the experience can be learned or adapted together with the target model training in an iterative way, as in Equation~\ref{eq:experience:advanced}. We show below that the discriminator/critic-based experience, in combination of certain choices of the divergence function $\BDist$, re-derives the  generative adversarial learning \citep{goodfellow2014generative} from a different perspective than Section~\ref{sec:divergence:gans}. 

{\bf Generative adversarial learning: The variational experience view.}
The functional descent view of generative adversarial learning presented in Section~\ref{sec:divergence:gans} is based on the treatment that the experience is the given static data instances, and the various GAN algorithms are due to the different choices of the divergence function. The extended view of SE in this section also allows an alternative viewpoint of the learning paradigm, that gives more flexibility in not only choosing the divergence function but also the experience function, leading to a richer set of GAN variants.

In this viewpoint, we consider experience that is defined variationally as mentioned above. That is, the experience function $f$, as a measure of the goodness of a sample $\bt$, is not specified a priori but rather defined through an optimization problem. As a concrete example, we define $f$ as a binary classifier $f_\phi$ with sigmoid activation and parameters $\bphi$, where the value $f_\phi(\bt)$ measures the {\it log} probability of the sample $\bt$ being a real instance (as opposed to a model generation). Thus the higher $f_\phi(\bt)$ value, the higher quality of sample $\bt$. The parameters $\bphi$ of the experience function need to be learned. We can do so by augmenting the standard equation (Equation~\ref{eq:se-loss}) with added optimization of $\bphi$ in various ways. The following equation gives one of the approaches:
\begin{equation}
\begin{split}
    \min_{q,\btheta} \max_{\bphi}&~~ - \alpha \BENT\left( q \right) + \beta \BDist\left( q, p_\theta \right) - \E_{q}\left[ f_\phi \right] + \E_{\pdata}\left[ f_{\phi} \right],
\end{split}
\label{eq:divg:var-gan-fgan-H}
\end{equation}
where, besides the optimization of $q$ and $\btheta$, we additionally maximize over $\bphi$ with the extra term $\E_{\pdata}\left[ f_\phi \right]$ to form the classification problem $\max_{\bphi} - \E_{q}\left[ f_\phi \right] + \E_{\pdata}\left[ f_{\phi} \right]$. Further assuming a particular configuration of the tradeoff hyperparameters $\alpha=0$ and $\beta=1$, the resulting objective
\begin{equation}
\begin{split}
    \min_{q,\btheta} \max_{\bphi}&~~ \BDist\left( q, p_\theta \right) - \E_{q}\left[ f_\phi \right] + \E_{\pdata}\left[ f_{\phi} \right],
\end{split}
\label{eq:divg:var-gan-fgan}
\end{equation}
turns out to relate closely to generative adversarial learning. 

In particular, with proofs adapted from \citet{farnia2018convex}, Equation~\ref{eq:divg:var-gan-fgan} recovers the vanilla GAN algorithm when $\BDist$ is the Jensen-Shannon divergence and assuming the space of $f_\phi$, denoted as $\mathcal{F}$, is convex.
More specifically, if we denote the probability $C_\phi(\bt) = \exp f_\phi(\bt)$, then the equation reduces to the familiar GAN objective in Equation~\ref{eq:divg:pfd-gan}. The results can be extended to the more general case of f-GAN~\citep{nowozin2016f}: 
if we set $\BDist$ to an $f$-divergence and do not restrict the form (e.g., classifier) of the experience function $f_\phi$, then with mild conditions, the equation recovers the f-GAN algorithm. Now consider $\BDist$ as the first-order Wasserstein distance and suppose the $f_\phi$-space $\mathcal{F}$ is a convex subset of 1-Lipschitz functions. It can be shown that Equation~\ref{eq:divg:var-gan-fgan} reduces to the Wasserstein GAN algorithm as shown in Equation~\ref{eq:divg:wgan} where $\varphi$ now corresponds to $f_\phi$. Note that for the above configurations, if $f_\phi$ is parameterized as a neural network with a fixed architecture (e.g., ConvNet), its space $\mathcal{F}$ is not necessarily convex (i.e., a linear combination of two neural networks in $\mathcal{F}$ is not necessarily in $\mathcal{F}$). In such cases we formulate the optimization of the experience function over $\mathrm{conv}(\mathcal{F})$, the convex hull of $\mathcal{F}$ containing any convex combination of neural network functions in $\mathcal{F}$ \citep{farnia2018convex}, and see the various GAN algorithms as approximations by considering only the subset $\mathcal{F}\subseteq\mathrm{conv}(\mathcal{F})$.

Besides the above examples of divergence $\BDist$ that each leads to a different GAN algorithm, we can consider even more options, such as the hybrid $f$-divergence and Wasserstein distance studied in \citep{farnia2018convex}. 
Of particular interest is to set $\BDist$ to the KL divergence $\BDist(q,p_\theta)=\KLD(q\|p_\theta)$, motivated by the simplicity in the sense that the auxiliary distribution $q$ has a closed-form solution:\footnote{We can alternatively derive the objective from Equation~\ref{eq:divg:var-gan-fgan-H}, by setting $\alpha=\beta=1, \BDist$ to the cross entropy, and $\BENT$ to the Shannon entropy. Note that $\KLD(q\|p_\theta)=-\ENT(q)-\E_{q}[\log p_\theta]$. Thus the solution of $q$ can be derived as in Equation~\ref{eq:se-teach-student}.} at each iteration $n$, 
\begin{equation}
\begin{split}
q^{(n+1)}(\bt) = p_{\theta^{(n)}}(\bt) \exp\left\{ f_{\phi^{(n)}}(\bt) \right\} ~/~ Z,
\end{split}
\label{eq:divg:var-q}
\end{equation}
where $Z$ is the normalization factor. 
As shown in \citet{wu2020improving}, the particular form of solution results in a new variant of GANs that enables more stable optimization of the experience function (discriminator) $f_\phi$. Concretely, the discriminator $f_\phi$ can now be optimized with importance reweighting:
\begin{equation}
\begin{split}
\max_{\bphi}& -\E_{\bt\sim q^{(n+1)}}\left[ f_\phi(\bt) \right] + \E_{\bt\sim p_d}\left[ f_\phi(\bt) \right] \\
=& -\frac{1}{Z}\E_{\bt\sim p_\theta^{(n)}}\Big[ \exp\{ f_{\phi^{(n)}}(\bt) \} \cdot f_{\phi}(\bt) \Big] + \E_{\bt\sim p_d}\left[ f_\phi(\bt) \right],
\end{split}
\label{eq:divg:var-d-weighting}
\end{equation}
where importance sampling is used to estimate the expectation under $q^{(n+1)}$, using the generator $p_\theta^{(n)}$ as the proposal distribution. Compared to the vanilla and Wasserstein GANs above, the fake samples from the generator are now weighted by the exponentiated discriminator score $\exp\{ f_{\phi^{(n)}}(\bt) \}$ when used to update the discriminator. Intuitively, the mechanism assigns higher weights to samples that can fool the discriminator better, while low-quality samples are downplayed to avoid degrading the discriminator performance during training.

Besides the generative adversarial learning, in Section~\ref{sec:experience} we briefly mentioned that many of the conventional experience can also benefit from the idea of introducing adaptive or learnable components, for example, data instances with automatically induced data weights or learned augmentation policies (Section~\ref{sec:experiences:data:manipulate}). We discuss more in Section~\ref{sec:app:repurposing} about how the unified SE as the basis can effortlessly derive efficient approaches for joint model and experience optimization.

\subsubsection{\textbf{Online Experience From the Environment}}\label{sec:dynamic:experience:online}

Another form of dynamic experience comes from the changing environments, such as the data stream in online learning \citep{shalev2012online} whose distribution can change over time, or the experience in lifelong learning \citep{thrun1998lifelong} that differs across a series of tasks.

In particular, we consider an online setting: at each time $\tau\in\{1, ..., T\}$, a predictor is given an input and is required to make a prediction (e.g., if the stock market will go up or down tomorrow where). We have access to the recommended prediction by each of the $K$ experts $\bt=\{1,...,K\}$, and make our prediction accordingly. As a result, the environment reveals a reward based on the discrepancy between the prediction and the true answer. The sequence of data instances follows a dynamic that is unknown and can even be adversarially adaptive to the predictor's behavior (e.g., in the problem of spam email filtering, or other strategic game environments) \citep{shalev2012online}. In such cases, we can only hope the predictor to achieve some relative performance guarantee, in particular w.r.t. the best single expert in hindsight. This is formally captured by {\it regret}, which is the difference between the cumulative reward of the predictor and that of the best single expert. 
We further consider the specific setting where we submit a `soft' prediction $p_\tau(\bt)$, that is, the distribution (a vector of normalized weights) over the $K$ experts, and receive the rewards $r_\tau(\bt)\in\mathbb{R}$ for each expert $\bt$. The goal of the learning problem is then to update the distribution $p_\tau(\bt)$ at every step $\tau$ for minimal regret $\sum\nolimits_{\tau=1}^{T} r_\tau(\bt^*) - \sum\nolimits_{\tau=1}^{T} \E_{p_\tau(\bt)}[r_\tau(\bt)]$, where $\bt^*$ is the best single expert.


The rewards from the environment naturally serves as the dynamic experience:
\begin{equation}
\begin{split}
    f_{\tau}(\bt) := r_{\tau}(\bt).
\end{split}
\label{eq:dynamic:online-experience}
\end{equation}
In the following, we show that the SE rediscovers the classical multiplicative weights (MW), or Hedge algorithm \citep{freund1997decision} to the above online problem. Similar ideas of multiplicative weights have been widely used in diverse fields such as optimization, game theory, and economics \citep{arora2012multiplicative}.

{\bf Multiplicative weights.}
To update the distribution (i.e., normalized weight vector) over the $K$ experts at each time $\tau$, we treat the distribution as the target model $p_\theta$ to be learned in the SE. In other words, $p_\theta$ is directly parameterized as the normalized weight vector $p_\theta := \btheta = \{\theta_{\bt}\}_{\bt=1}^{K}$, where $\theta_{\bt}\geq0$ is the probability of expert $\bt$, with $\sum_{\bt=1}^K \theta_{\bt}=1$. Starting from the SE in Equation~\ref{eq:se-loss}, and assuming $\BDist$ to be the cross entropy, $\BENT$ the Shannon entropy, and $\alpha=\beta>0$, we obtain the update rule of the target model at each time $\tau$ following the teacher-student mechanism in Equation~\ref{eq:se-teach-student}. Specifically, the teacher step has:
\begin{equation}
\begin{split}
\text{Teacher:}\quad &q^{(\tau+1)}(\bt) = p_{\theta^{(\tau)}}(\bt) \exp\left\{ \alpha^{-1} f_\tau(\bt) \right\} ~/~ Z. 
\end{split}
\label{eq:dynamic:mw-teacher}
\end{equation}
The subsequent student step is to minimize the cross entropy between the target model $p_\theta$ and $q^{(\tau+1)}$ (see Equation~\ref{eq:se-teach-student}). Given the definition of $p_\theta$ as the vector of probabilities, the student step is equivalent to directly setting $p_\theta$ to the vector of $q^{(\tau+1)}(\bt)$ values. Therefore:
\begin{equation}
\begin{split}
\text{Student:}\quad &p_{\theta^{(\tau+1)}}(\bt) = p_{\theta^{(\tau)}}(\bt) \exp\left\{ \alpha^{-1} f_\tau(\bt) \right\} ~/~ Z,
\end{split}
\label{eq:dynamic:mw-student}
\end{equation}
which is precisely the multiplicative weight update rule for the expert distribution/weights. That is, at each time, the weights are updated by multiplying them with a factor that depends on the reward.

\subsection{Dynamic SE as Interpolation of Algorithms}\label{sec:dynamic:other}
Besides the dynamic experience, other components in the SE, such as the divergence function $\BDist$ and the balancing weights $(\alpha, \beta)$, can also be indexed by the time $\tau$ with desired evolution. In particular, the previous sections have shown that the different specifications of the SE components correspond to different specific algorithms, many of which are well-known and have different properties. The dynamic SE with the evolving specifications, therefore, can be seen as \emph{interpolating} between the algorithms during the course of training. The interpolation allows the learning to enjoy different desired properties in different training stages, resulting in improved efficacy.

As a concrete example, \citet{tan2018connecting} learn a text generation model by starting with the simple supervised MLE algorithm, with the experience function $f_{\tau=0}=f_{\text{data}}$ (Equation~\ref{eq:experience:data-sup}) and balancing weights $(\alpha_{\tau=0}=1, \beta_{\tau=0}=\epsilon)$ as described in Section~\ref{sec:experience:data:sup}. After warming up with the supervised MLE for $n$ iterations, the approach then changes the experience function to the data augmentation-based one $f_{\tau=n}=f_{\text{data-aug}}$ defined in Equation~\ref{eq:experience:augment}, which introduces noise and task-specific evaluation information into the training. As revealed in Section~\ref{sec:experiences:data:manipulate}, this stage in effect corresponds to the reward-augmented maximum likelihood (RAML) algorithm \citep{norouzi2016reward}. The approach proceeds by further annealing the balancing weights, specifically by gradually increasing $\beta_\tau$ from $\epsilon$ to $1$ as $\tau$ increases. The increase of $\beta$ effectively gets the learning closer to a reinforcement-style learning (notice that the policy gradient algorithm has $\alpha=\beta=1$ as described in Section~\ref{sec:experience:reward:Q}). Intuitively, the target model $p_{\theta^{(t)}}$, as part of the $q^{(\tau+1)}$ solution in the teacher step weighted by the increasing weight $\beta$ (see Equation~\ref{eq:se-teach-student}), serves to produce more data samples. Those samples are weighted by the experience function and used for updating the target model further, simulating the policy gradients. Therefore, the whole learning procedure spans multiple paradigms of learning (from supervised MLE, RAML, to reinforcement learning) that increasingly introduces more noise and exploration for improved robustness. All the involved paradigms of algorithms are perfectly encompassed in the single SE formulation, allowing users to simply tweak and evolve the relevant components for the interpolation.

\section{Optimization Algorithms}\label{sec:opt}

Thus far, we have discussed the standard equation as the unified objective function. Learning the target model $p_\theta$ amounts to optimizing the objective w.r.t the model parameters $\btheta$. That is, the standardized objective presents an optimization problem, for which an optimization solver is applied to obtain the target model solution $p_{\theta^*}$. This section is devoted to discussion of various optimization algorithms.

For a simple objective such as that of the vanilla supervised MLE (Equation~\ref{eq:mle-ori}) with a tractable model, stochastic gradient descent can be used to optimize the model parameters $\btheta$ straightforwardly. With a more complex model
or a more complex objective like the general SE, more sophisticated optimization algorithms are needed, such as the teacher-student procedure exemplified in Equation~\ref{eq:se-teach-student}.
Like the standardized formulation of the objective function, we would like to quest for a \emph{standardized optimization algorithm} that is generally applicable to optimizing the objective under vastly different specifications.
Yet it seems still unclear whether such a universal solver exists and what it looks like. On the other hand, some techniques may hold the promise to generalize to broad settings.


\begin{figure}
    \centering
    \includegraphics[width=0.5\textwidth]{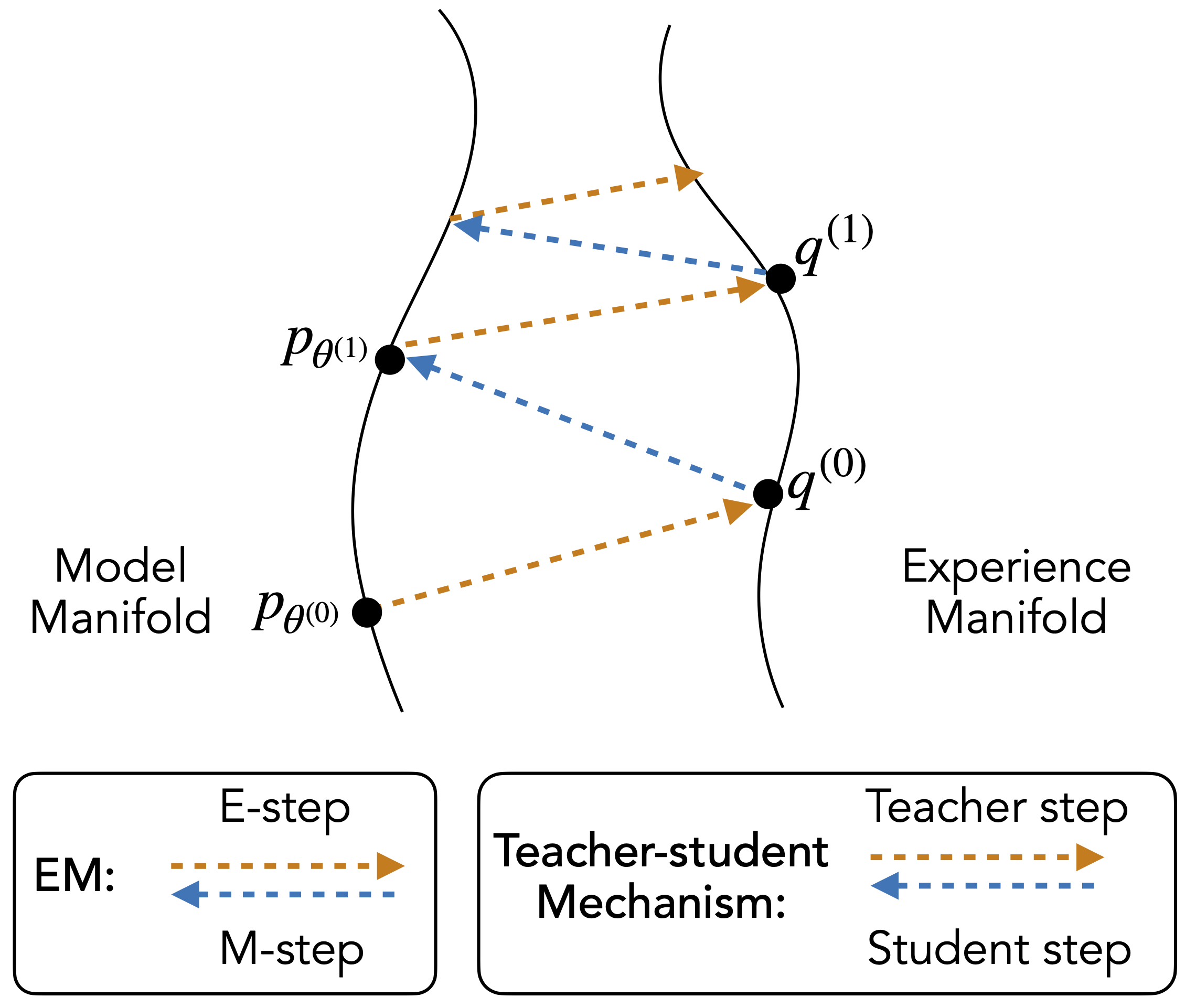}
    \caption{EM and the teacher-student mechanism entail alternating projection to achieve local minima.}
    \label{fig:opt:alternating-projection}
\end{figure}

\subsection{Alternating Projection}\label{sec:opt:alternating}
Alternating projection \citep{csiszar1975divergence,bauschke1996projection} provides a general class of optimization algorithms from a geometry point of view, subsuming as special cases many of the optimization algorithms discussed above, ranging from the
EM (Section~\ref{sec:unsup-mle}) to the teacher-student algorithm (Section~\ref{sec:se}). At a high level, the algorithms consider the optimization problem as to find a common point of a collection of sets, and achieve it by projecting between those sets in succession. 
For example, 
the EM algorithm entails alternating projection, as shown in Figure~\ref{fig:opt:alternating-projection}, where the E-step (Equation~\ref{eq:em-e}) projects the current model distribution $p_{\theta^{(n)}}(\x,\y)$ onto the set of distributions whose marginal over $\x$ equals the empirical distribution, i.e., $q^{(n+1)} = \argmin_q \KLD\left( q(\y|\x)\tilde{p}_d(\x) \| p_{\theta^{(n)}}(\x,\y)  \right)$, then the subsequent M-step (Equation~\ref{eq:em-m}) projects $q^{(n+1)}$ onto the set of possible model distributions through $\min_\theta \KLD\left( q^{(n+1)}(\y|\x)\tilde{p}_d(\x) \| p_\theta(\x,\y) \right)$.

In general, the SE in Equation~\ref{eq:se} or \eqref{eq:se-loss}, with both the model parameters $\btheta$ and the auxiliary distribution $q$ to be learned, can naturally be optimized with an alternating-projection style procedure, which we have referred to as the teacher-student mechanism. 

\subsection{The Teacher-Student Mechanism}

We have seen a special case of the teacher-student mechanism in Equation~\ref{eq:se-teach-student} for solving the specifically instantiated SE (e.g., with cross entropy as the divergence function $\BDist$). The optimization procedure is also an example of alternating projection (Figure~\ref{fig:opt:alternating-projection}). Specifically, 
the teacher $q^{(n+1)}$ is the projection of the student $p_{\theta^{(n)}}$ onto the set defined by the experience, and the student $p_{\theta^{(n+1)}}$ is the projection of the teacher $q^{(n+1)}$ onto the set of model distributions. 

\subsubsection{\textbf{The Teacher Step}}\label{sec:opt:teacher}

The teacher step in Equation~\ref{eq:se-teach-student} has a closed-form solution for the teacher $q^{(n+1)}$ due to the choice of cross entropy as the divergence function $\BDist$ in SE:
\begin{equation}
\begin{split}
\text{Teacher:}\quad &q^{(n+1)}(\bt) = \exp\left\{ \frac{\beta \log p_{\theta^{(n)}}(\bt) + f(\bt)}{\alpha} \right\} ~/~ Z.
\end{split}
\label{eq:opt-se-teach}
\end{equation}
In the more general case where other complex divergence functions are used (as those in Section~\ref{sec:divergence}), a closed-form teacher is usually not available. 
The probabilistic functional descent (PFD) mentioned in Section~\ref{sec:divergence:gans}, with approximations to the influence function using convex duality, offers a possible way of solving for $q$ for a broader class of divergences, such as Jensen-Shannon divergence and Wasserstein distance. It thus presents a promising venue for future research to develop more generic solvers for the broad learning problems characterized by SE. 

On the other hand, as can be seen in the student step discussed shortly, sometimes we do not necessarily need a closed-form teacher $q^{(n+1)}$ in the learning, but only need to be able to draw samples from $q^{(n+1)}$.

{\bf Probability functional descent.}
Generally, the SE (Equation~\ref{eq:se} or \ref{eq:se-loss}) defines a loss over the auxiliary distribution $q$, denoted as $J(q)$, which is a functional on the auxiliary distribution space $\mathcal{Q}(\tspace)$. The G\^{a}teaux derivative of $J$ at $q$, if exists, is defined as \citep{fernholz2012mises}:
\begin{equation}
\begin{split}
 J'_{q}(h-q) = \lim_{\epsilon\to 0^{+}}\frac{J(q+\epsilon(h-q))-J(q)}{\epsilon}
\end{split}
\label{eq:divg:g-derivative}
\end{equation}
for any given $h\in\mathcal{Q}(\tspace)$. Intuitively, $J'_{q}(h-q)$ describes the change of the $J(q)$ value with respect to an infinitesimal change in $q$ in the direction of $(h-q)$. The G\^{a}teaux derivative $J'_{q}(h-q)$ can alternatively be computed with the \emph{influence function} of $J$ at $q$, denoted as $\psi_q: \tspace\to\mathbb{R}$, through:
\begin{equation}
\begin{split}
    J'_{q}(h-q) = \int_{\bt} \psi_q(\bt) (h-q) d\bt = \E_{h}\left[ \psi_q(\bt) \right] - \E_{q}\left[ \psi_q(\bt) \right].
\end{split}
\label{eq:divg:inf-func}
\end{equation}
The above notions allow us to define gradient descent applied to the functional $J$. Concretely, we can define a linear approximation to $J(q)$ around a given $q_0$:
\begin{equation}
\begin{split}
    J(q) &\approx J(q_0) + J'_{q_0}(q-q_0) \\
    &= J(q_0) + \E_{q}\left[ \psi_{q_0}(\bt) \right] - \E_{q_0}\left[ \psi_{q_0}(\bt) \right] \\
    &= \E_{q}\left[ \psi_{q_0}(\bt) \right] + const.
\end{split}
\label{eq:divg:expansion}
\end{equation}
Thus, $J(q)$ can approximately be minimized with an iterative descent procedure: at each iteration $n$, we perform a descent step that decreases $\E_{q}\left[ \psi_{q^{(n)}}(\bt) \right]$ w.r.t. $q$, yielding $q^{(n+1)}$ for the next iteration.

Once the functional gradient is defined as above, the remaining problem of the optimization is then about how to obtain the influence function $\psi_q$ given the functional $J(q)$. In some cases the influence function as defined in Equation~\ref{eq:divg:inf-func} is not directly tractable and approximations are needed. \citet{chu2019probability} developed a variational approximation method applied when $J$ is convex (which is the case in Equation~\ref{eq:se-loss} when $\BDist$ is convex w.r.t $q$). Concretely, with the convex conjugate of $J$ defined as $J^*(\varphi) = \sup\nolimits_{h} \E_{h} \left[ \varphi(\bt) \right] - J(h)$, it can be shown under mild conditions that the influence function for $J$ at $q$ is:
\begin{equation}
\begin{split}
    \psi_q = \argmax\nolimits_{\varphi\in\mathcal{C}(\tspace)}  \E_{q}\left[ \varphi(\bt) \right] - J^*(\varphi),
\end{split}
\label{eq:divg:inf-func-convex-dual}
\end{equation}
where $\mathcal{C}(\tspace)$ is the space of continuous functions $\tspace\to\mathbb{R}$. We thus can approximate the influence function by parameterizing it as a neural network and training the network to maximize the objective $\E_{q}\left[ \varphi(\bt) \right] - J^*(\varphi)$. Plugging the approximation of influence function into the above functional descent procedure leads to the full PFD optimization:
\begin{equation}
\begin{split}
    \inf_{q} \sup_{\varphi} \E_{q}\left[ \varphi(\bt) \right] - J^*(\varphi),
\end{split}
\label{eq:divg:pfd-adv}
\end{equation}
which is a saddle-point problem.

\subsubsection{\textbf{The Student Step}}\label{sec:opt:student}

The student step optimizes the SE objective w.r.t. the target model parameters $\btheta$, given $q^{(n+1)}$ from the teacher step. The optimization is to minimize the divergence between the student $p_\theta$ and the teacher $q^{(n+1)}$:
\begin{equation}
\begin{split}
    \text{Student:}\quad &\bm{\theta}^{(n+1)} = \argmin_{\btheta} \BDist\left( q^{(n+1)}, p_\theta \right).
\end{split}
\label{eq:opt-se-student}
\end{equation}
Section~\ref{sec:divergence} discussed the different choices of the divergence function $\BDist$. In particular, in the same setting of Equation~\ref{eq:se-teach-student} where $\BDist$ is the common cross entropy (or KL divergence as discussed in Section~\ref{sec:divergence:ce}), the student step is written as:
\begin{equation}
\begin{split}
    \bm{\theta}^{(n+1)} = \argmax_{\btheta} \E_{q^{(n+1)}(\bt)} \big[ \log p_\theta(\bt) \big],
\end{split}
\label{eq:opt-se-student-ce}
\end{equation}
where $q^{(n+1)}$ is in the form of Equation~\ref{eq:opt-se-teach} above. The optimization then amounts to first drawing samples from the teacher $\tilde{\bt} \sim q^{(n+1)}$, and then updating $\btheta$ by maximizing the log-likelihood of those samples under $p_\theta$. We could apply various sampling methods to draw samples from $q^{(n+1)}$, such as Markov chain Monte Carlo (MCMC) methods, like Gibbs sampling when $\bt$ is discrete and Hamiltonian or Langevin MC for continuous $\bt$ \citep{duane1987hybrid,neal1992bayesian,welling2011bayesian}, and sampling based on stochastic differential equations \citep[SDEs,][]{song2020score,liu2022composable}.

In the special case of $\alpha=\beta$, the teacher model reduces to $q^{(n+1)}(\bt) \propto p_{\theta^{(n)}}(\bt) \exp\{ f(\bt) / \alpha \}$. This special form allows us to use a simple {\it importance sampling} method \citep{hu2018deep}, with $p_{\theta^{(n)}}$ (i.e., the current model distribution) as the proposal distribution:
\begin{equation}
\begin{split}
    \bm{\theta}^{(n+1)} = \argmax_{\btheta} \E_{\tilde{\bt} \sim p_{\theta^{(n)}}(\bt)} \big[ \exp\{ f(\tilde{\bt}) / \alpha \} \cdot \log p_\theta(\tilde{\bt}) \big] ~/~ Z.
\end{split}
\label{eq:opt-se-student-importance-sampling}
\end{equation}
Here the experience function $f(\bt)$ serves as a sample reweighting mechanism \citep{wu2020improving} that highlights the `high-quality' samples (i.e., those receiving high goodness scores in the light of the experience) for updating the parameters $\btheta$. 

For other choices of the divergence function $\BDist$ other than the cross entropy or KL divergence, sampling from $q^{(n+1)}$ may similarly be sufficient in order to estimate and optimize the divergence Equation~\ref{eq:opt-se-student}.  The probability functional descent (PFD) can also be used to optimize with a certain class of divergences, following  similar derivations discussed in Sections~\ref{sec:divergence} and \ref{sec:opt:teacher}.

\section{The Target Model}\label{sec:target-model}

Besides the  objective function (Sections~\ref{sec:se}-\ref{sec:advanced-experiences}) and the optimization (Section~\ref{sec:opt}) above, we now briefly discuss the third and last core ingredient of an ML approach, namely, the target model $p_\theta$. A key feature of the SE as an objective function is that the formulation is agnostic to the specific choices of the target model. That is, one can use the SE to train effectively anything with a learnable component (denoted as $\btheta$ in general), ranging from black-box neural networks, to probabilistic graphical models, symbolic models, and the combinations thereof. 
We will mention a few examples of the target model that are of radically different categories, illustrating the utility and generality of the SE.

\begin{figure}
    \centering
    \includegraphics[width=\textwidth]{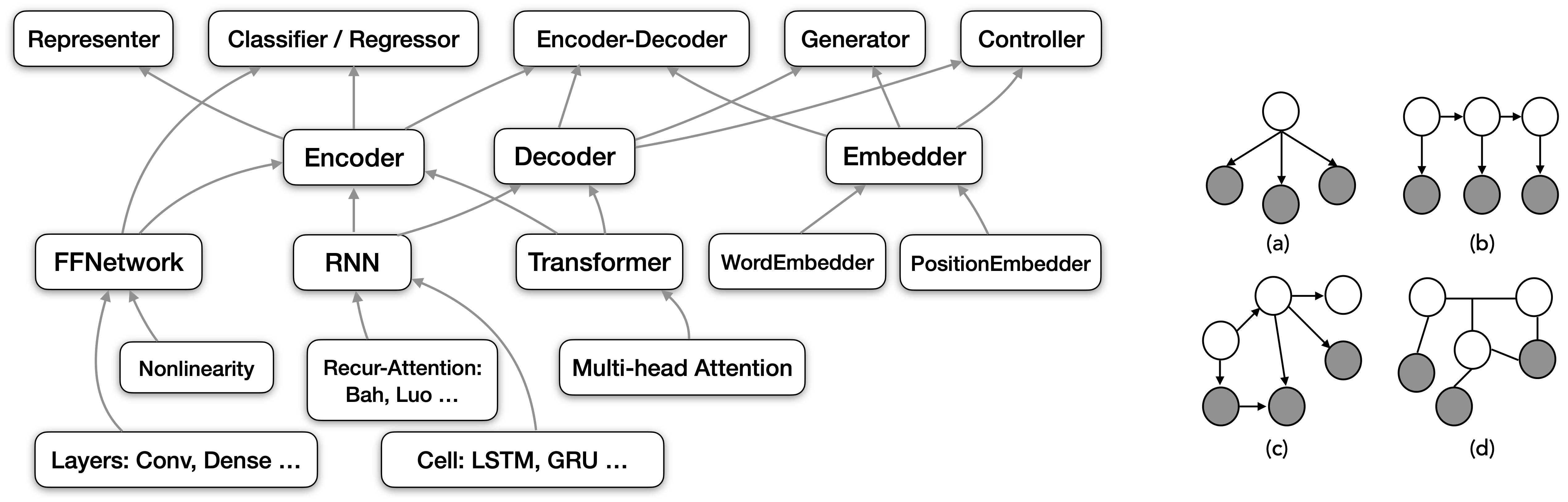}
    \vspace{-10pt}
    \caption{{\bf Left:} An (incomplete) set of neural model components at different granularities. Components at a lower level are composed to form higher-level, more complex ones. {\bf Right:} Directed/undirected probabilistic graphical models of various dependence structures.}
    \label{fig:model:arch}
\end{figure}

{\bf Deep neural networks of any architectures.}
The target model can be a deep neural network of an arbitrary architecture, composed of different neural layers and units. Figure~\ref{fig:model:arch}, right panel, illustrates a set of common neural components and how they are composed to form more complex ones at different levels of granularity. Neural models of different architectures are used for different tasks of interest. For example, an image classifier $p_\theta(\y|\x)$, with input image $\x$ and output object label $\y$, typically consists of an encoder and a classification layer (a.k.a., classification head), where the encoder can be a ConvNet \citep{lecun1989backpropagation} or transformer \citep{vaswani2017attention} and the classification layer is often a simple multilayer perceptron (MLP) consisting of several neural dense layers. As another example, a neural language model $p_\theta(\bt)$ over text sequence $\bt$ is a generator with a decoder component (which is in turn implemented via a transformer or other neural architectures).

The $\btheta$ to be learned includes the neural network weights. It can be randomly initialized and then trained from scratch. On the other hand, with the prevalence of \emph{pretrained models}, such as BERT \citep{devlin2019bert} for text representation and GPT-3 \citep{brown2020language} for language modeling, we can often initialize $\btheta$ with the appropriate pretrained model weights and finetune it to the downstream tasks of interest. Sometimes it is sufficient to finetune only a subset of the model parameters (e.g., the classification head of the image classifier) while keeping all other parts fixed.
In many downstream tasks, it is often difficult to obtain a large number of supervised data instances to train/finetune $\btheta$ with simple supervised MLE. In such cases,  SE  offers a more flexible framework that allows plugging in all other forms of experience (as those in Section~\ref{sec:experience}) related to the downstream tasks for more effective learning.

\begin{figure}
    \centering
    \includegraphics[width=\textwidth]{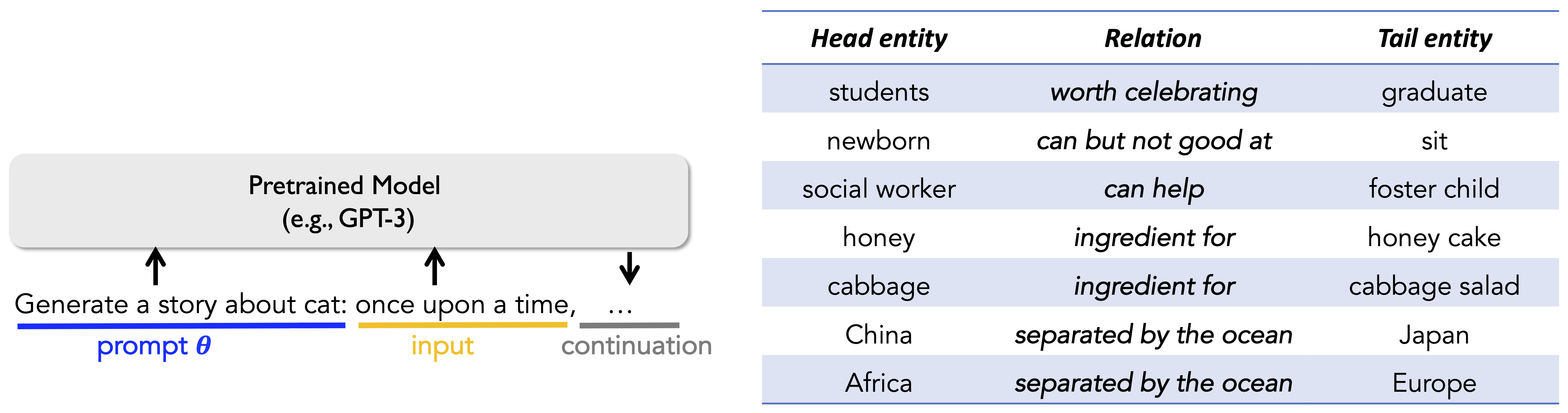}
    \vspace{-10pt}
    \caption{
    {\bf Left:} prompt as the learnable component $\btheta$ to steer pretrained models (such as GPT-3) to perform the downstream task of interest (e.g., generating a story of a certain topic). {\bf Right:} The novel learned knowledge tuples by treating the symbolic knowledge graph as the target model \citep{hao2022bertnet}.
    }
    \label{fig:model:prompt-kg}
\end{figure}

{\bf Prompts for pretrained models.}
Updating large pretrained models can be prohibitively expensive due to the massive amount of model parameters. Prompting is a new emerging way of steering the pretrained models for performing downstream tasks without changing the model parameters \citep{brown2020language}. A prompt is a short sequence of text tokens or embedding vectors which, by concatenating with the input and being fed together into the model, stimulates the model to perform the task of interest and produce the desired output for the input (Figure~\ref{fig:model:prompt-kg}, left). It is thus desirable to learn the  prompt that enables optimal performance of the pretrained models  on the task of interest. For continuous prompt, which is a sequence of differentiable embedding vectors \citep{li2021prefix,lester2021power}, we could naturally treat the prompt as the learnable $\btheta$, and plug the resulting target model $p_{\theta}$, now the pretrained model plus the learnable prompt, into the SE for training. On the other hand, for discrete prompt, which is a sequence of text tokens, one can instead parameterize a prompt-generation network as $\btheta$ which, after training, generates optimal discrete prompts for the downstream task. For example, \citet{deng2022rlprompt} used both supervised data instances and task reward to learn the target prompt-generation models.

{\bf Symbolic knowledge graphs.}
The target model can even be a symbolic system such as a knowledge graph (KG) or a rule set. Here $\btheta$ denotes the KG structure to be learned, and $p_\theta(\bt)$ can be seen as a distribution assigning a (non-)uniform nonzero probability to any knowledge tuple $\bt$ in the KG and zero to all other tuples. \citet{hao2022bertnet} presents a concrete instance of the learning system that incrementally learns (extracts) a commonsense relational KG (Figure~\ref{fig:model:prompt-kg}, right), using the pretrained language models such as BERT \citep{devlin2019bert} as the experience.

{\bf Probabilistic graphical models and composite models.}
The target model $p_\theta(\bt)$ can also be probabilistic graphical models \citep{jordan2003introduction,koller2009probabilistic}, a rich family of models characterizing the conditional dependence structure between random variables with a directed/undirected graph (Figure~\ref{fig:model:arch}, right). Graphical models may also be composed with the neural modules to form more complex composite models, typically with the neural modules extracting features from the raw inputs and the graphical modules capturing the high-level structures \citep{wilson2016deep,johnson2016composing,zheng2015conditional}. As discussed in the earlier sections, the common learning and inference approaches for probabilistic graphic models, such as the (variational) EM algorithm, are special instances of SE and its teacher-student mechanism. The SE framework offers a generalized formulation for learning graphical and composite models.

\section{Panoramic Learning with All Experience}\label{sec:app}

The preceding sections have presented a standardized formalism of machine learning, on the basis of the standard equation of objective function, that provides a succinct, structured formulation of
a broad design space of learning algorithms, 
and subsumes a wide range of known algorithms in a unified manner. The simplicity, modularity, and generality of the framework is particularly appealing not only from the theoretical perspective but also because it offers guiding principles for mechanical design of algorithmic approaches to challenging problems, in the presence of diverse experience, and hence serves as an important step toward the goal of panoramic learning.
In this section, we discuss the use of the standard equation to drive systematic design of new learning methods, which in turn yield various algorithmic approaches to problems in different application domains.

\subsection{Combining Rich Experience}\label{sec:app:combine}

As one of the original motivations for the standardization, the framework allows us to combine together all different experience to learn models of interest. Learning with multiple types of experience is a necessary capability of AI agents to be deployed in a real dynamic environment and to improve from heterogeneous signals in different forms, at varying granularities, from diverse sources, and with different intents (e.g., adversaries). Such versatile learning capability is best exemplified by human learning---for instance, we learn a new language by reading and hearing examples, studying grammars and rules, practicing and interacting with others, receiving feedback, associating with prior languages we already mastered, and so forth.
To build a similar panoramic-learning AI agent, having a standardized learning formalism is perhaps an indispensable step.

The standard equation we have presented is naturally designed to fit the needs. In particular, the experience function provides a straightforward vehicle for experience combination. For example, the simplest approach is perhaps to make a weighted composition of multiple individual experience functions:
\begin{equation}
\begin{split}
    f(\bt) = \sum\nolimits_{i} \lambda_i f_i(\bt),
\end{split}
\label{eq:learn-all-experience-combine}
\end{equation}
where each $f_i$ characterizes a specific source of experience, and $\lambda_i>0$ is the respective weight. One can readily plug in arbitrary available experience, such as data, logical rules, constraints, rewards, and auxiliary models, as components of the experience function. 

With the SE, designing an approach to a problem thus boils down to choosing and formulating {\it what} experience to use depending on the problem structure and available resources, without worrying too much about {\it how} to use the experience. This provides a potentially new modularized and repeatable way of producing ML approaches to different problems, as compared to the previous practice of designing bespoke algorithm for each individual problem.
Sections~\ref{sec:experience} and \ref{sec:advanced-experiences} have discussed possible formulations of the diverse types of experience as an experience function to be plugged into Equation~\ref{eq:learn-all-experience-combine}. It is still an open question how even more types of experience, such as massive knowledge graphs \citep{tan2020summarizing,bach2019snorkel}, can efficiently be formulated as an experience function that assesses the `goodness' of an input $\bt$. On the other hand, the discussion in the next subsection offers new opportunities that relieve users from having to manually specify every detail of the experience function. Instead, users only need to specify  parts of the experience function of which they are certain, and leave the remaining parts plus the weights $\lambda_i$ (Equation~\ref{eq:learn-all-experience-combine}) to be automatically learned together with the target model. 


\begin{table}[t]
    \centering
    \small
    \begin{tabular}{@{}l l@{}}
    \toprule
        Method & Negative to Positive \\\midrule
        Original & {it was super {\color{blue} dry} and had a {\color{blue} weird taste} to the entire slice .} \\
        \citep{shen2017style} & {it was super {\color{red} friendly} and had a {\color{red} nice touch} to the same .} \\
        $f_{\text{sc}}$ & {\color{red} good good good} \\
        $f_{\text{sc}} + f_{\text{data}}$ &  it was super {\color{red} well-made} and had a {\color{blue} weird taste} to the entire slice . \\
        $f_{\text{sc}} + f_{\text{data}} + f_{\text{LM}}$ & {it was super {\color{red} fresh} and had a {\color{red} delicious taste} to the entire slice .} \\ [0.1cm]
        \toprule
        Method & Positive to Negative \\\midrule
        Original & besides that , the wine selection they have is {\color{red} pretty awesome} as well . \\
        \citep{shen2017style} &  after that , the quality prices that {\color{blue} does n’t pretty much well} as . \\
        $f_{\text{sc}}$ & besides {\color{blue} horrible  horrible} as  \\
        $f_{\text{sc}} + f_{\text{data}}$ &  besides that , the wine selection they have is {\color{blue} pretty borderline as atrocious} . \\
        $f_{\text{sc}} + f_{\text{data}} + f_{\text{LM}}$ &  besides that , the wine selection they have is {\color{blue} pretty horrible} as well . \\
    \bottomrule
    \end{tabular}
    \begin{tabular}{l c c c}
\cmidrule[\heavyrulewidth](lr){1-4}
 \multirow{1}{*}{Method} & \multirow{1}{*}{Attribute accuracy ($\uparrow$)}  & \multicolumn{1}{c}{Preservation ($\uparrow$)} & \multirow{1}{*}{Fluency ($\downarrow$)}  \\
\cmidrule[\heavyrulewidth](lr){1-4}
\citep{shen2017style} & 79.5\% & 12.4 & 51.4 \\
\cmidrule(lr){1-4}
$f_{\text{sc}}$ & 99.6\% & 1.2 & 259.2  \\
$f_{\text{sc}} + f_{\text{data}}$ & 87.7\% & 65.6 & 177.7 \\ 
$f_{\text{sc}} + f_{\text{data}} + f_{\text{LM}}$ & 91.2\% & 57.8 & 53.95 \\
\cmidrule[\heavyrulewidth](lr){1-4} 
\end{tabular}
    \caption{Results of text attribute transfer on the Yelp data \citep{hu2017controllable,yang2018unsupervised}. {\bf Top: } Generated samples of different methods given the {\it Original} sentence, for negative-to-positive and positive-to-negative transfer, respectively. Positive and negative words are highlighted in {\color{red} red} and {\color{blue} blue}, respectively.
    {\bf Bottom: }Performance of different methods. For the evaluation metrics: ``attribute accuracy'' is evaluated with a pretrained sentiment classifier on the model generations; ``preservation'' is the BLEU score ($\in [0, 100]$) that evaluates the lexical overlapping between the generated sentences and the original input sentences; ``fluency'' is the perplexity score of generated text evaluated by a pretrained language model (LM) (the lower, the better). The model trained with all three types of experience achieves the best overall performance in terms of all the three aspects.}
    \label{tab:tst-results}
\end{table}

{\bf Case study: Text attribute transfer.} 
As a case study of learning from rich experience, consider the problem of text attribute transfer where we want to rewrite a given piece of text to possess a desired attribute \citep{hu2017controllable,shen2017style}. 
Taking the sentiment attribute, for example, given a sentence $\x$ (e.g., a customer's review {\it ``the manager is a horrible person''}) and a target sentiment $a$ (e.g., positive), the goal of the problem is to generate a new sentence $\y$ that (1) possesses the target sentiment, (2) preserves all other characteristics of the original sentence, and (3) is fluent (e.g., the transferred sentence ``{\it the manager is a perfect person}''). To learn an attribute transfer model $p_\theta(\y|\x,a)$, a key challenge of the problem is the lack of direct supervision data (i.e., pairs of sentences that are exact the same except for sentiment), making it necessary to use other forms of experience. 
Here we briefly describe an approach originally presented in \citet{hu2017controllable,yang2018unsupervised}, highlighting how the approach can be built mechanically, by formulating relevant experience directly based on the problem definition and then plugging them into the SE. 

We can identify three types of experience, corresponding to the above three desiderata, respectively. First, the model needs to learn the concept of `sentiment' to be able to modify the attribute of text. 
As mentioned in Section~\ref{sec:experinece:model}, a natural form of experience that has encoded the concept is a pretrained sentiment classifier (SC). 
The first experience function can then be defined as $f_{\text{sc}}(\x,a,\y)=\text{SC}(a,\y)$, which evaluates the log likelihood of the transferred sentence $\y$ possessing the sentiment $a$. The higher value the $\y$ achieves, the higher quality it is considered in light of the experience. 
The second desideratum requires the model to reconstruct as much of the input text $\x$ as possible. We combine the second experience $f_{\text{data}}(\x,a,\y|\dataset)$ (Equation~\ref{eq:experience:data-sup}) defined by a set of simple reconstruction data instances $\dataset = \{(\x^*, a^*=a_{\x^*}, \y^*=\x^*)\}$, where the target sentiment $a^*$ is set to the sentiment of the original sentence $\x^*$, and by the problem definition the ground-truth output $\y^*$ is exact the same as the input $\x^*$. Such data thus carry the information of preserving the input content.
Finally, for the requirement of generating fluent text, we can again naturally use an auxiliary model as the experience, namely, a pretrained language model (LM) $f_{\text{LM}}(\x,a,\y)=\text{LM}(\y)$ that estimates the log likelihood of a sentence $\y$ under the natural language distribution. 
After identifying the experience $(f_{\text{sc}}, f_{\text{data}}, f_{\text{LM}})$, we then combine them together with Equation~\ref{eq:learn-all-experience-combine} and plug into the SE to train the target model $p_\theta(\y|\x,a)$. More experimental details can be found in \citet{hu2017controllable,yang2018unsupervised}. 
Table~\ref{tab:tst-results} shows the empirical results on the common Yelp corpus of customer reviews. We can see that by plugging in the relevant experience, the resulting model successfully learns the respective aspects of the task. For example, with only $f_{\text{sc}}$, the model is able to transfer the sentiment attribute but fails on content preservation and fluency. Adding the second experience $f_{\text{data}}$ encourages preservation. Further with $f_{\text{LM}}$, the model substantially improves the fluency, achieving the best overall performance. The case study demonstrates the necessity of integrating the diverse experience for solving the problem.










\subsection{Repurposing Learning Algorithms for New Problems}\label{sec:app:repurposing}

The standardized formalism sheds new light on fundamental relationships between a number of learning problems in different research areas, showing that they are essentially the same under the SE perspective.
This opens up a wide range of opportunities for generalizing existing algorithms, which were originally designed for specialized problems, to a much broader set of new problems. It is also made easy to exchange between the diverse research areas in aspects of modeling, theoretical understanding, approximation, and optimization. 
For example, an earlier successful approach to challenges in one area can now be readily applied to address challenges in another. Similarly, a future progress made in one problem could immediately unlock progresses in many others.


\begin{figure}
    \centering
    \includegraphics[width=0.8\textwidth]{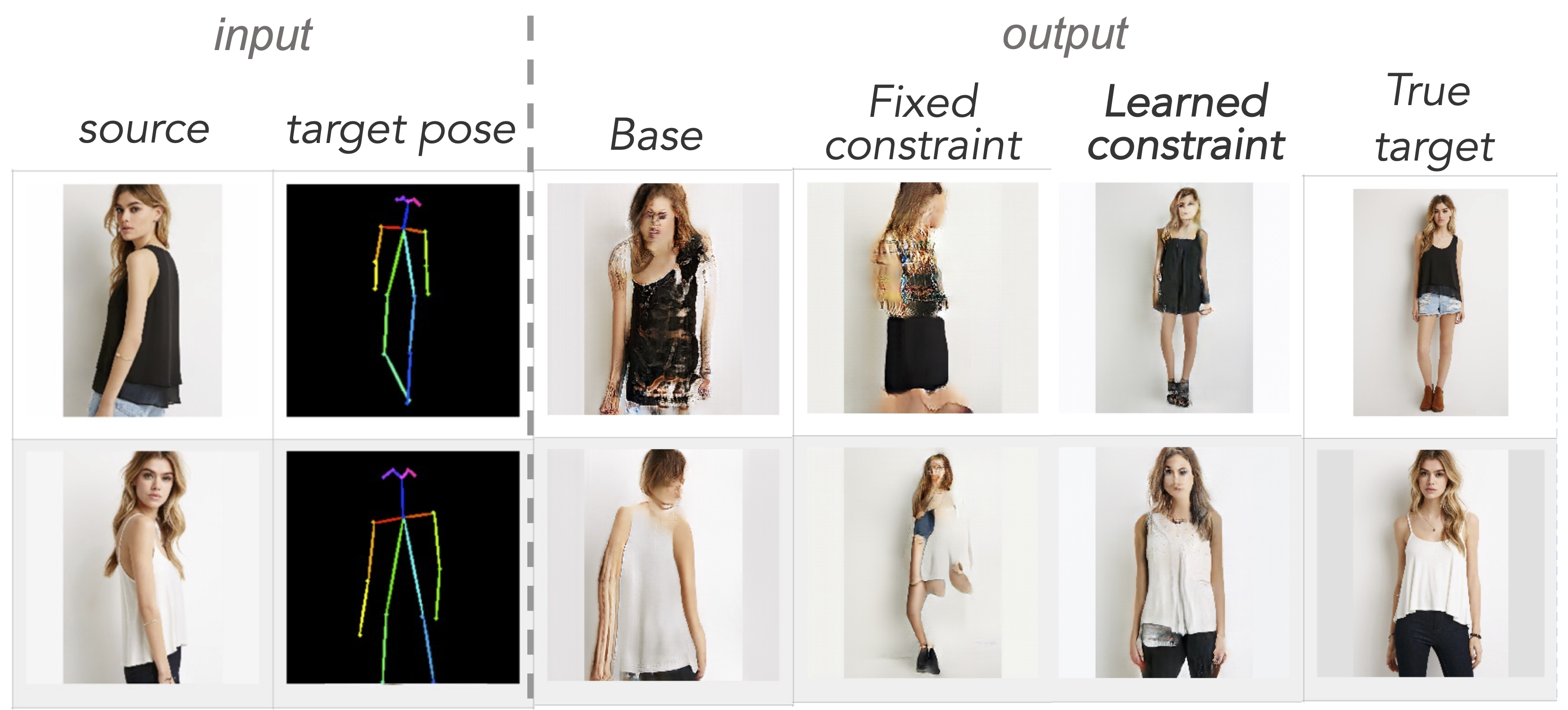}
    \caption{Example outputs of pose-conditioning human image generation. Given an image of a person as well as a target pose represented as a skeleton, the goal is to generate a new image of the same person under the target pose. The {\it base} model that learns a neural model with only limited available supervised data fails to make meaningful generation. \citet{hu2018deep} added a structured constraint on the human body structure, based on a pretrained human part parser. Learning with the {\it fixed constraint} (i.e., with the pretrained parser) does not improve the generation quality. In contrast, learning the constraint together with the target model within the dynamic standard equation framework leads to substantially improved output ({\it learned constraint}), close to the {\it true target}.}
    \label{fig:case:human-pose}
\end{figure}

{\bf Case study: Learning with imperfect experience.}
To illustrate, let us consider a set of concrete problems concerning with distinct types of experience, which were often studied by researchers in different areas: 
{\bf (1)} The first problem is to integrate structured knowledge constraints in model training, where some components of the constraints, as well as the constraint weights, cannot be specified a priori and are to be induced automatically; {\bf (2)} The second problem concerns supervised learning, where one has access to only a small set of data instances with imbalanced labels, and we want to automate the data manipulation (e.g., augmentation and reweighting) to maximize the training performance; {\bf (3)} The last problem is to stabilize the notoriously difficult training of generative adversarial networks (GANs) for a wide range of image and text generation tasks. 

The three problems, though seemingly unrelated at first sight, can all be reduced to the same underlying problem in the unified SE view, namely, learning with imperfect experience $f$ (e.g., underspecified knowledge constraints, small imbalanced data, and unstable discriminator). We want to automatically adapt/improve the imperfect experience
in order to better supervise the target model training. 
This readily falls into the dynamic SE setting described in Section~\ref{sec:advanced-experiences},
where now the experience function is $f_\phi(\bt)$ associated with learnable parameters $\bphi$. 
For example, in problem (1), $f_\phi(\bt)=\sum_i\lambda^i_\phi f^i_{\text{rule}, \phi}(\bt)$, where any learnable components in each knowledge constraint $f^i_{\text{rule}, \phi}$ (Section~\ref{sec:experience:knowledge}) and the weights $\lambda^i_\phi$ constitute the $\bphi$ to be learned \citep{hu2018deep}. In problem (2), $f_\phi(\bt)$ is instantiated as $f_{\text{data-w},\phi}(\bt;\dataset)$ (Equation~\ref{eq:experience:reweighting}) with learnable  
data weights $w(\bt^*) \in \bphi$, or as $f_{\text{data-aug},\phi}(\bt;\dataset)$ (Equation~\ref{eq:experience:augment}) with the metric for augmentation $a_{\bt^*}(\bt) \in \bphi$ to be learned \citep{hu2019learning}. In problem (3), we have discussed the training of $f_\phi(\bt)$ as the GAN discriminator in Section~\ref{sec:divergence:gans:var}, but we want to improve the training stability \citep{wu2020improving}. Thus, one approach for efficient updates of the general experience function $f_\phi$ would address all three problems together. 

To seek for solutions, we again take advantage of the unified SE perspective that enables us to reuse existing successful techniques instead of having to invent new ones. In particular, the connection of the experience function $f$ with the reward in Section~\ref{sec:experience:reward} naturally inspires us to repurpose known techniques from the fertile reinforcement learning (RL) literature, especially those of learning reward functions such as inverse RL \citep{ziebart2008maximum} or learning implicit reward \citep{zheng2018learning}, for learning the experience function $f_\phi$ in our problems. For instance, following \citep{ziebart2008maximum}, one can acquire and update the experience function at each iteration in Equation~\ref{eq:experience:advanced} through $\min_{\bphi} -\E_{\bt^* \sim \tilde{p}_{\text{data}}}\left[ \log q(\bt^*) \right]$,
where $q$ taking the form in Equation~\ref{eq:se-teach-student} now depends on $\bphi$. The resulting procedure induces an importance reweighting scheme that is shown to stabilize the discriminator training in GANs \citep{wu2020improving}, as well as learn meaningful constraints \citep{hu2018deep}. Figure~\ref{fig:case:human-pose} provides a demonstration that learning the constraints together with the target model within the dynamic SE framework leads to substantial improvement.

\section{Related Work}\label{sec:related}

It has been a constant aspiration to search for basic principles that unify the different paradigms in machine learning \citep{langley1989toward,bishop2013model,domingos2015master,gori2017machine,hu2019workshop}. Extensive efforts have been made to build unifying views of methods on particular fronts. For example, \citet{roweis1999unifying} unified various unsupervised learning algorithms with a linear Gaussian model; \citet{wainwright200511} presented the variational method for inference in general exponential-family graphical models; \citet{richardson2006markov,domingos2015master} presented Markov logic networks that combine Bayesian Markov network with first-order logic for uncertain inference; \citet{knoblauch2019generalized} developed a generalized form of variational Bayesian inference by allowing losses and divergences beyond the standard likelihood and KL divergence, which subsumes existing variants for Bayesian posterior approximation; \citet{altun2006unifying} showed the duality between regularized divergence (e.g.,  Bregman and $f$-divergence) minimization and statistical inference (e.g., MAP estimation); \citet{arora2012multiplicative} presented a common formulation of different multiplicative weights update methods; \citet{mohamed2016learning} connected generative adversarial learning with a rich set of other statistical learning principles;
\citet{wu2019tale} discussed connections between generative, discriminative, and energy-based models. \citet{ma2022principles} presented parsimony and self-consistency as the guiding principles for learning from data. 
Those unified treatments shed new light on the sets of originally specialized methods and foster new progress in the respective fields. 
\citet{lecun2022path} presented a modeling architecture to construct autonomous intelligent agents that combines concepts such as world model and hierarchical joint embedding. Our standardized formalism of the learning objective is complementary and offers a general framework for training the relevant model architectures. The framework also covers the key learning ingredients mentioned in \citet{lecun2022path}, including the self-supervised learning (Section~\ref{sec:experience:data:selfsup}) and intrinsic motivation (Section~\ref{sec:experience:reward:intrinsic}).

Integrating diverse sources of information in training has been explored in previous work, which is often dedicated to specific tasks. \citet{roth2017incidental} presented different ways of deriving supervision signals in different scenarios. \citet{zhu2020dark} discussed the integration of physical and other knowledge in solving vision problems. The distant or weak supervision approaches \citep{mintz2009distant,ratner2017snorkel} automatically create (noisy) instance labels from heuristics, which are then used in the supervised training procedure. The panoramic learning we discussed here makes use of broader forms of experience not necessarily amenable to be converted into supervised labels, such as reward, discriminator-like models, and many structured constraints. The experience function $f(\y)$ offers such flexibility for expressing all those experiences.

\section{Future Directions}\label{sec:future}

We have presented a standardized machine learning formalism, materialized as the standard equation of the objective function, that formulates a vast algorithmic space governed by a few components regarding the experience, model fitness measured with divergence, and uncertainty. The formalism gives a holistic view of the diverse landscape of learning paradigms, allows a mechanical way of designing ML approaches to new problems, and provides a vehicle toward panoramic learning that integrates all available experience in building an AI agent. The work shapes a range of exciting open questions and opportunities for future study. We discuss a few of these directions below.

{\bf Continual learning in complex dynamic environments.}
We have discussed the SE for learning with all diverse forms of experience, in both static environments (e.g., fixed data or reward distributions) and dynamic environments (e.g., optimized or online experience). An exciting next step is to deploy the SE framework to build an AI agent that continually learns in the real-world complex and fast-evolving context, in which the AI agent must learn to identify the relevant experience out of massive external information, to acquire increasingly complex new concepts or skills. 
Establishing and applying the standardized formalism to the broader learning settings is expected to unleash even more power by enabling principled design of learning systems that continuously improve by interacting with and collecting diverse signals from the outer world.


{\bf Theoretical analysis of panoramic learning.}
The paradigm of panoramic learning poses new questions about theoretical understanding. A question of particular importance in practice is about how we can guarantee better performance after integrating more experience. The analysis is challenging because the different types of experience can each encode different information, sometimes noisy and even conflicting with each other (e.g., not all data instances would comply with a logic rule), and thus plugging in an additional source of experience does not necessarily lead to positive effects. But before that, a more basic question to ask is, perhaps, how can we characterize learning with some special or novel forms of experience, such as logic rules and auxiliary models? What mathematical tools may we use for characterization, and what would the convergence guarantees, complexity, robustness, and other theoretical and statistical properties be?
Inspired by how we generalized the specialized algorithms to new problems, a promising way of the theoretical analysis would be to again leverage the standard equation and repurpose the existing analyses originally dealing with supervised learning, online learning, and reinforcement learning, to now analyze the learning process with all other experiences. 

{\bf From standardization to automation.}
As in other mature engineering disciplines such as mechanical engineering, standardization is followed by automation. The standardized ML formalism opens up the possibility of automating the process of creating and improving ML algorithms and solutions. The current `AutoML' practice has largely focused on automatic search of neural network architectures and hyperparameters, thanks to the well-defined architectural and hyperparameter search spaces. 
The standard equation that defines a structured algorithmic space would similarly suggest opportunities for automatic search or optimization of learning algorithms, which is expected to be more efficient than direct search on the programming code space~\citep{real2020automl}.
We briefly discussed in Section~\ref{sec:app} how new algorithms can mechanically be created by composing experience and/or other algorithmic components together. It would be significant to have an automated engine that further streamlines the process. For example, once a new advance is made to reinforcement learning, the engine would automatically amplify the progress and deliver enhanced functionalities of learning data manipulation and adapting knowledge constraints. Similarly, domain experts can simply input a variety of experiences available into their own problems, and expect an algorithm to be automatically composed to learn the target model they want. The sophisticated algorithm manipulation and creation would greatly simplify machine learning workflow in practice and boost the accessibility of ML to much broader users.

From Maxwell's equations to General Relativity, and to quantum mechanics and Standard Model, ``Physics is the study of symmetry,'' remarked physicist Phil Anderson \citep[][p.394]{anderson1972more}. The end goal of physics research seems to be clear---a `theory of everything' that fully explains and links together all physical aspects. The `end goal' of ML/AI is surely much more elusive. Yet the unifying way of thinking would be incredibly valuable, to continuously unleash the extensive power of current vibrant research, to produce more principled understanding, and to build more versatile AI solutions.

\printbibliography



\end{document}